\documentclass[accepted]{uai2021} % after acceptance, for a revised
\usepackage[american]{babel}
\usepackage{natbib} % has a nice set of citation styles and commands
\usepackage{mathtools} % amsmath with fixes and additions
\usepackage{booktabs} % commands to create good-looking tables
\usepackage{tikz} % nice language for creating drawings and diagrams

\title{Bayesian optimization for modular black-box systems with switching costs}

\usepackage{color}
\definecolor{ggreen}{rgb}{0,0.79,0}
\definecolor{orange}{rgb}{0.79,0.375,0.08}

\newcommand{\alg}{LaMBO}

\usepackage{algorithm}
\usepackage{algorithmic}
\usepackage{amsmath}
\usepackage{amssymb}
\usepackage{amsthm}
\usepackage{wrapfig}
\usepackage{comment}
\usepackage{graphicx}
\usepackage{bbm}

\definecolor{mydarkblue}{rgb}{0,0.08,0.45}
\hypersetup{ %
pdftitle={},
pdfauthor={},
pdfkeywords={},
pdfborder=0 0 0,
pdfpagemode=UseNone,
colorlinks=true,
linkcolor=mydarkblue,
citecolor=mydarkblue,
filecolor=mydarkblue,
urlcolor=black,
pdfview=FitH}

\usepackage{mathtools}

\usepackage{graphicx}

\makeatletter
\newtheorem*{rep@theorem}{\rep@title}
\newcommand{\newreptheorem}[2]{%
\newenvironment{rep#1}[1]{%
 \def\rep@title{#2 \ref{##1}}%
 \begin{rep@theorem}}%
 {\end{rep@theorem}}}
\makeatother

\newtheorem{lemma}{Lemma}

\newtheorem{theorem}{Theorem}
\newreptheorem{theorem}{Theorem}

\newtheorem{remark}{Remark}
\newreptheorem{lemma}{Lemma}

\author[1]{\href{mailto:<clin354@gatech.edu>?Subject=Bayesian optimization for modular black-box systems with switching costs}{Chi-Heng Lin}{}}
\author[2]{Joseph D. Miano}
\author[1,3]{Eva L. Dyer}
\affil[1]{%
    School of Electrical and Computer Engineering\\
    Georgia Tech\\
    Atlanta, Georgia, USA
}
\affil[2]{
    School of Computer Science\\
    Georgia Tech\\
    Atlanta, Georgia, USA
}
\affil[3]{%
    Department of Biomedical Engineering\\
    Georgia Tech \& Emory University\\
    Atlanta, Georgia, USA
}

\begin{document}

\maketitle

\begin{abstract}
Most existing black-box optimization methods assume that all variables in the system being optimized have equal cost and can change freely at each iteration. However, in many real-world systems, inputs are passed through a sequence of different operations or {\em modules}, making variables in earlier stages of processing more costly to update. Such structure induces a dynamic cost from {\em switching} variables in the early parts of a data processing pipeline. In this work, we propose a new algorithm for switch-cost-aware optimization called Lazy Modular Bayesian Optimization (\alg). This method efficiently identifies the global optimum while minimizing cost through a passive change of variables in early modules.
The method is theoretically grounded which achieves a vanishing regret regularized with switching cost. We apply \alg~to multiple synthetic functions and a three-stage image segmentation pipeline used in a neuroimaging task, where we obtain promising improvements over existing cost-aware Bayesian optimization algorithms. Our results demonstrate that \alg~is an effective strategy for black-box optimization capable of minimizing switching costs.
\end{abstract}

\section{Introduction}\label{sec:intro}
Bayesian optimization (BO) \citep{ snoek2012practical,Srinivas:2010:GPO:3104322.3104451,mockus1978application} is a popular technique that is used to optimize unknown black-box systems. Such systems arise in a wide range of applications ranging from robotics \citep{berkenkamp2016safe} and sensor networks \citep{10.1145/1791212.1791238}, to hyperparameter tuning in machine learning \citep{10.5555/2986459.2986743,frazier2018tutorial}. In the black-box setting, the underlying function that maps variables to a reward (loss) is unknown and is instead queried. 
%Through probabilistic modeling with stochastic processes, BO approximates the unknown function with a Gaussian process. 
BO methods find ways to tackle this challenging setting by approximating the unknown function with a Gaussian process (GP) \citep{rasmussen2003gaussian} and updating this belief on the fly to decide which sample to generate next. %BO methods have emerged as a popular class of approaches for black-box optimization.
%Leveraging potential structure in the functions that we are interested in optimizing is critical for having any hope in tackling this problem. \eva{some last sentence that says how GPs help us}

%B BO increases the accuracy of the approximation and quality of optimization on the fly.

Unfortunately, when trying to optimize a complex black-box system, the cost of generating a sample can often be prohibitive. Here, costs could represent the amount of time, energy, or resources required to generate a black-box sample (i.e., test a new hyperparameter parameter configuration of interest). To account for costs to update different variables, or overall cost constraints, a wide range of different cost-aware and multi-resolution sampling strategies ranging from batch optimization \citep{gonzalez2016batch,kathuria2016batched}, multi-fidelity model \citep{kandasamy2016gaussian,kandasamy2017multi}, multi-objective optimization \citep{abdolshah2019cost}, to dynamic programming \citep{lam2017lookahead,lam2016bayesian} have been developed over the past decade.

While the underlying black-box function that we want to optimize may be unknown, many real-world systems have costs with specific structure that are known ahead of time. An important yet simple abstraction of many systems encountered in practice is that they process their inputs through a sequence of {\em modules}, where the outputs from one module to the next are chained together. For instance, in many scientific applications like genomics \citep{davis2017genomics} and neuroimaging \citep{abraham2014machine,johnson2019toward}, generating an output (sample) often involves running high-dimensional inputs through multiple stages (modules) of processing, and each module has unique hyperparameters that must be optimized. When making updates in these types of sequential systems,
it becomes much more costly to update a variable at an earlier stage of processing because we must take into account the fact that all operations in subsequent stages must be rerun. Not only does this sequential structure affect the cost, but it also gives rise to \textit{switching costs}, where the cost depends on which variables are modified between consecutive iterations. However, most of these methods are agnostic to additional information about the structure of the underlying costs in the system, and thus are too aggressive in changing variables across modules.

%we can model the cost of updating a variable as the amount of accumulated cost that we require to update a variable at a specific stage of the processing chain. All of the remaining variables at later stages are allowed to move freely without adding any further cost. This means that at a subsequent iteration, the cost of changing a variable, depends on the action you select or which stage of processing you want to go back to, and thus will also vary dynamically. 

%When optimizing these systems in an end-to-end manner, the cost to not only query, but also to switch to new variable settings at early stages in a pipeline, can be prohibitive. 

%while cost-aware methods have been explored in a number of contexts, the dynamic nature of \text{switching costs} has not yet been well addressed with existing approaches. Most literature assumes the cost to be static with respect to previous decisions, however, there are many cases where costs switch over time and depending on the information and actions selected.

In light of these motivations, we introduce a new algorithm for black-box optimization called Lazy Modular Bayesian Optimization (\alg). This method leverages modular and sequential structure in a system to reduce overall cumulative costs during optimization. To quantify the cost of switching in these cases, we model the cost of each query as the aggregation of cost needed to rerun modules from the first step where a variable must be updated. 
%In this scenario, when variables at later stages of processing are updated, the outputs from earlier modules are frozen (stored) and can be used to facilitate downstream optimization until its necessary to switch variables earlier on. This idea can be codified with a new notion of  {\em movement regret} for Bayesian optimization, which measures both the functional optimality and cost of changing variables in the system. 
By encouraging the optimization method to be lazy, analytically we show that \alg~achieves a sublinear rate in a notion of switching-cost regularized regret. %To the best of our knowledge, \alg~is the first algorithm with strong theoretical guarantees that incorporates this type of system structure into Bayesian optimization. 
 %\alg~combines a Slowly Moving Bandit (SMB) \citep{NIPS2017_7000,pmlr-v65-koren17a} strategy with  Gaussian Process (GP) inference to provide a cost-aware strategy for black-box optimization.
We also empirically evaluate the performance of the proposed method by applying \alg~to a number of synthetic datasets and neuroimaging problem where the aim is to tune a modular pipeline for 3D reconstruction of neuroanatomical structures from slices of 2D images \citep{lee2019convolutional, johnson2019toward}.  
%When compared with traditional Bayesian optimization (BO) baselines \citep{Srinivas:2010:GPO:3104322.3104451,movckus1975bayesian,wang2017max} and cost-aware variants \citep{snoek2012practical,lee2020costaware,abdolshah2019cost}, we find that our method outperforms these methods in terms of the trade-off between deviation from global optimum and the cumulative cost.  We further apply \alg~to a problem arising in neuroimaging where the aim is to produce a 3D segmentation of brain structure \citep{lee2019convolutional, johnson2019toward}.  
%In this application, we are tasked with end-to-end optimization of a three-module pipeline for 3D reconstruction of neuroanatomical structures from slices of 2D images. 
%The three modules correspond to three sequential operations: data pre-processing, pixel-level semantic segmentation with a deep neural network (Unet) \citep{ronneberger2015u}, and data post-processing steps to form a 3D reconstruction. 
Our empirical results show that hyperparameters in this three-stage system can be optimized to $95\%$ optimality jointly over multiple modules within 1.4 hours compared with 5.6 hours obtained from the best of the alternatives. These results point to the fact that leveraging system structure, and dynamic switching costs, can be advantageous for optimizing multi-stage black-box systems.  

\vspace{-2mm}
\paragraph{Summary of Contributions.}
\label{sec:contribution}
The contributions of this work are as follows:
({\em i})~In Section \ref{sec:Approach}, we 
formulate a novel Bayesian optimization problem with \textit{switching-cost} constraints, and propose the algorithm \alg~to solve the problem in systems with modular structure. 
To the best of our knowledge, this is the first attempt to leverage modular system structure in the design of a cost-efficient algorithm for black-box optimization.
({\em ii})~In Section \ref{sec:analysis}, we establish theoretical guarantees of \alg~ by proving a regularized regret bound taking switching-cost consumption into consideration using techniques from both the multi-armed bandit and Bayesian optimization literature.
({\em iii})~In Section \ref{sec:exp}, we apply our method to synthetic functions and to a 3D brain-image-segmentation task. We empirically demonstrate that the method can efficiently solve switch-cost-aware optimization across modular compositions of functions. 
%({\em iv})~In Section \ref{sec:exp}, we apply our method to a 3D brain image segmentation task
%, where the processing steps are represented in a sequential-block structure. We 
%and show that by minimizing variable switching in early modules, we can optimize performance while also reducing the total cost needed.

\section{Background and Related Work}\label{sec:Background}

\subsection{Bayesian Optimization}
\label{sec:BO}
 Black-box optimization methods aim to find the global minimum of an unknown function $f(x)$ with only a few queries. Let $f^*$ and $\mathbf{x}^*$ be the optimal function value and optimizer, respectively. Standard algorithms seek to produce a sequence of inputs $\mathbf{x}^1,\dots,\mathbf{x}^T$ that result in (potentially) noisy observations $\mathbf{y}^1,\dots,\mathbf{y}^T$ such that $f(\mathbf{x}^t)$ will approach the optimal value $f^*$ quickly. A common choice to measure performance of a candidate algorithm is the \textit{cumulative regret}: 
 \begin{equation}
 R(T) = \sum_{t=1}^Tf(\mathbf{x}^t)-f^*.
  \end{equation}
Among the many different approaches for black-box optimization, BO is a celebrated probabilistic method whose statistical inferences are tractable and theoretically grounded. It uses a Gaussian process (GP) prior on the distribution of the unknown function $f$, which is characterized by a mean function $\mu( {\bf x} )$ and a kernel function $k_0( {\bf x} , {\bf x}')$. Let ${\bf k}_t(\mathbf{x}):=[k_0(\mathbf{x},\mathbf{x}^1),\dots,k_0(\mathbf{x},\mathbf{x}^t)]^T$,  $\mathbf{K}_t:=[k_0(\mathbf{x}^i,\mathbf{x}^j)]_{1\leq i,j\leq t}$, and $\sigma^2$ represent the noise variance.  In this case, we can  update the posterior with simple closed-form formulas: 
\begin{align}\label{postupdate}
    \mu_{t+1}(\mathbf{x})&=\mathbf{k}_t^T(\mathbf{x})(\mathbf{K}_t+\sigma \mathbf{I})^{-1}\mathbf{y}_t,\nonumber\\
    \sigma_{t+1}^2(\mathbf{x})&=k_0(\mathbf{x},\mathbf{x})-\mathbf{k}_t^T(\mathbf{x})(\mathbf{K}_t+\sigma \mathbf{I})^{-1}\mathbf{k}_t(\mathbf{x}).
\end{align}

Common classes of selection algorithms that use a BO framework include the: Upper Confidence Bound (UCB) \citep{Srinivas:2010:GPO:3104322.3104451}, Expected-Improvement (EI) \citep{10.1007/BFb0006170}, and entropy search \citep{wang2017max} algorithms.
At the heart of all of these methods is the design of an acquisition function that is used to select the next evaluation point, i.e., ${\bf x}^t \in \arg\min_{\bf x} \alpha^t({\bf x})$. The acquisition function allows flexibility in trading-off between exploration and exploitation and are constructed using the posterior statistics.  In this paper, we will adopt the UCB acquisition function due to its simplicity and success in both theory and practice. 
The GP-UCB acquisition function is given by 
\begin{equation}
    \alpha^t_{UCB}(\mathbf{x}) = \mu_{t-1}(\mathbf{x}) - \beta_t\sigma_{t-1}(\mathbf{x}),
    \end{equation}
where $\beta_t$ is a design parameter that controls the amount of exploration in the algorithm.
%\vspace{-1mm}
%\subsection{Cost-aware Bayesian Optimization}\label{sec:relatedWork}
%\vspace{-1mm}

\subsection{Slowly Moving Bandit Algorithm} \label{subsec:smb1}
To incorporate switching costs into a BO sampling strategy, we adopt \citep{pmlr-v65-koren17a} on solving a multi-armed bandit problem with switching costs. In this setting, optimization is formulated into a arm-selection problem where optimal variables $i$ (arms) are selected from a set $\cal{K}$ to minimize an unknown loss function $\ell :\cal{K}\mapsto \mathbb{R}$. At each iteration $t$, we can query an oracle to measure the loss (inverse reward) $\ell({i^t})$ by pulling arm $i^t$. In the switch-cost-aware case, there is a cost metric $c$ which incurs cost $c(i^t,i^{t-1})$ when switching between arms from $t-1$ to $t$. The objective is to minimize a linear combination of the loss and switching cost. In \citep{pmlr-v65-koren17a}, the authors propose the slowly moving bandit algorithm (SMB) to tackle the problem with a general cost metric. Here, we extend the idea to the setting of black-box optimization. 

%Because this method will be employed in our later algorithm, we will describe it in more detail. 
SMB is based on a multiplicative update strategy \citep{auer2002finite} that encodes the cost of switching between arms in a tree; each arm is a leaf and the cost to switch from one arm to another is encoded in the distance from their corresponding leafs in the tree. At each iteration $t$, SMB chooses an arm according to a probability distribution $p_t$ conditioned on the level of the tree (the root is level $0$) selected at the last iteration. We will make the sampling distribution precise momentarily. The distribution is then updated with a standard multiplicative update rule $p_t \leftarrow  p_t \exp({-\eta \widetilde{{\ell}}_t})$, where $\eta$ is the learning rate and $\tilde{\ell}_t$ is the estimated loss. Compared with basic bandit algorithms, there are two key modifications in SMB. First, it uses conditional sampling
to encourage slow switching. This constrains the arm selection to be the close to the previous choice, where distance is embedded in the tree's structure. Formally,  an arm is drawn according to the following conditional distribution $p(\cdot|A_{h_{t-1}}(i^{t-1}))$, where $h_{t-1}$ is a random level chosen at previous iteration, and $A_{h}(i)$ denotes the leaves (arms) that belong to the subtree rooted at level $h$ which has $i$ as one of its leaves. This ensures that $i^t$ remains in some small subtree as in the previous iteration.
Second, to utilize the classic multiplicative method, 
SMB makes sure that in average the conditional sampling is equivalent to direct sampling by modifying the loss estimators ${\tilde{\ell}}_{t}$ as,
\begin{align}\label{recursiveloss}
   \tilde{{\ell}}_t&=\bar{{\ell}}_{t,0}+\sum_{h=0}^{H-1}\sigma_{t,h}{\bar{\ell}}_{t,h}, \\ {\bar{\ell}}_{t,h}(i)&=\log\left(\sum_{j\in A_h(i)}\frac{p_t(j)e^{-\eta(1+\sigma_{t,h-1}){\bar{\ell}}_{t,h-1}(j)}}{p_t(A_h(i))}\right)^{-\frac{1}{\eta}},
\end{align}
where $\bar{\ell}_{t,0}$ is an unmodified loss estimator for algorithms without switching cost, and $\{\sigma_{t,k}\}_k$ are i.i.d. uniform random variables in $\{-1,1\}$.
For the purpose of self-contained, we include the pseudo-code of SMB in Supp. \ref{supp:smb}.

\vspace{-1mm}
\subsection{Related work}
The closest framework to ours in Bayesian optimization is the cost-aware Bayesian optimization, where instead of trying to minimize a function using the fewest samples, the methods strive to find the optimizer with least cumulative cost. The most standard method \citep{snoek2012practical,lee2020costaware} measures the acquisition function in the unit of the cost $\alpha^t(\mathbf{x})/c(\mathbf{x})^{\gamma}$, where $c$ denotes the cost function and $\gamma$ is some trade-off parameter. Another approach is to impose explicit cumulative budget constraints \citep{lam2016bayesian,lam2017lookahead}, where the authors have used dynamic programming-based approaches. While many of these algorithms have proposed cost-efficient optimization strategies under static costs, the scenario with the \textit{switching} cost where deviating from a previous action induces larger costs, has not been well-understood in the literature.
%Multi-objective BO \citep{abdolshah2019cost} by defining preferences over variables affected by the cost. \eva{preferences over variables?} 

Multi-fidelity strategies
\citep{kandasamy2017multi,poloczek2017multi,wu2020practical,mcleod2017practical} are also popular choices in which the decision maker is allowed to choose an additional fidelity parameter that controls the accuracy and the cost for function evaluations. In the sense that across subsequent resolutions there are correlations or structure in the costs of different parameters. On the surface, it seems our problem can be easily cast under the framework. However, as the dependencies on the accuracy of function approximation to variables in different modules are non-separable, one can not map a module to a fidelity. 
Another cost efficient BO approach similar to our work is process constrained BO \citep{vellanki2017process}, where some variables are not allowed to change due to constraints from physical system. %In this scenario, the authors propose efficient algorithm with nested BO utilizing parallel optimization strategy. 
CA-MOBO \citep{abdolshah2019cost} is also a cost-aware strategy which uses the framework of multi-objective optimization.  
It generalizes the UCB method to multi-dimensional outputs seeking a sweet spot in the trade-off between cost consumption and optimization accuracy.
Our work differs from theirs as we are allowed to probe variables anytime but may incur different cost when changing sets of variables in different modules, and we consider costs arising from switching variables. 

The switching cost optimization has been studied in multi-armed bandit literature \citep{kalai2005efficient,NIPS2017_7000,pmlr-v65-koren17a,dekel2014bandits,feldman2016online}. However, the arms are assumed to be uncorrelated, while in our work we assume strong dependency and leverage it by using Gaussian surrogate to explore multi-arms simultaneously.
\vspace{-1mm}

\section{Lazy Modular Bayesian Optimization (LaMBO)}
\label{sec:Approach}

\begin{figure}[!t]
  \centering
    \includegraphics[width=0.48\textwidth]{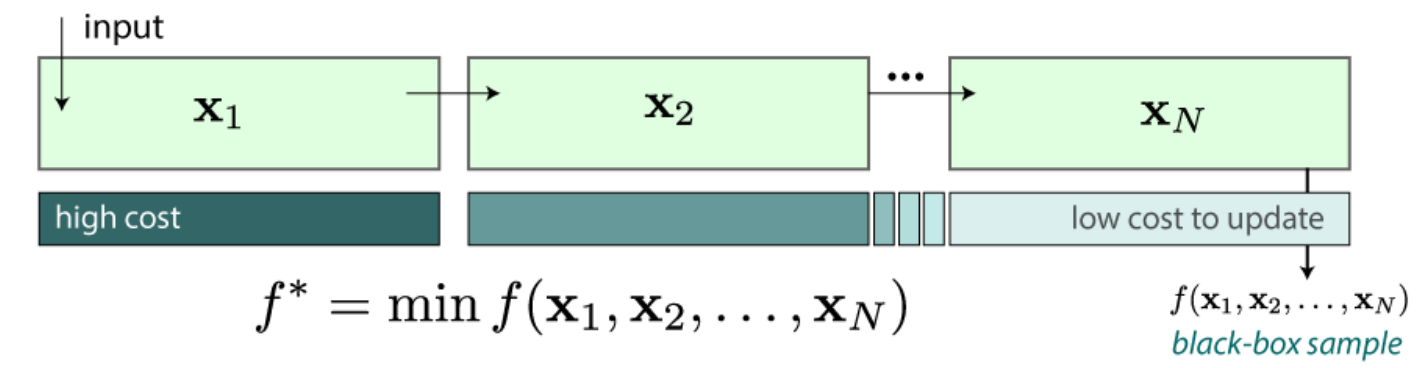}
    \caption{\footnotesize {\em Example of a modular system that consists of a sequence of operations that are applied, each with their own distinct set of variables}. When variables in early stages are changed, all the remaining modules need to run and this incurs high costs.}
    \vspace{-4mm}
    \label{fig:mobo1}
\end{figure}

A key assumption underlying this work is that the black-box system of interest has a modular structure, where the overarching system can be decomposed into a sequence of different sets of operations, each with a distinct set of  variables that need to be optimized (Figure~\ref{fig:mobo1}). 

\subsection{Problem Setup}
\label{sec:ProSetup}
%Such systems are encountered in robotics \citep{chu2018real}, image and signal analysis pipelines in the sciences \citep{johnson2019toward}, and in genomics applications \citep{davis2017genomics}. 
%The model comprises of $N$ modules represented by $N$ consecutive blocks. 
Let $\mathbf{x}_m\in {\cal{X}}_m$ denote the variables in the $m^{\rm th}$ module, and let $\mathbf{x}\in{\cal{X}}={\cal{X}}_1\times {\cal{X}}_2 \times\dots\times {\cal{X}}_N$ denote the set of  variables across all modules.
Our main goal is to propose a cost-efficient algorithm that finds the optimizer for a black-box function,
$$\mathbf{x}^*\in \arg\min_{\mathbf{x}\in\mathcal{X}} f(x).$$ 
The function $f$ is unknown to us, but when a set of variables $\mathbf{x} $ are input into the system,  this generates a noisy output $\mathbf{y}= f(\mathbf{x}) + \epsilon$, where $\epsilon$ is $\sigma$-sub-Gaussian.
To ensure that our model of the cost reflects the modular structure of the system, we make the following assumptions: (i) running  the $m^{\rm th}$ module incurs $c_m$ cost $,\forall m=1,\dots,N$, (ii) a module needs to be run only if variables in some modules earlier than it in the pipeline has been changed from previous iteration. We will also assume that
$c_N$, as any update requires updating variables in the last module, is negligible and 
equal to $0$.
Under the above modeling assumptions, the total cost incurred at iteration $t$ is equal to, \begin{equation}
\label{eq:movementcost}
\Gamma^t:=\sum_{m=1}^{N-1}c_m\mathbbm{1}_{\{\mathbf{x}^t_{1:m}\neq \mathbf{x}^{t-1}_{1:m}\}},
\end{equation}
where $\mathbbm{1}_{\{\mathbf{x}^t_{1:m}\neq \mathbf{x}^{t-1}_{1:m}\}}$ is an indicator that equals to $1$ when any variable in modules before the $m^{\rm th}$ module have been changed from the previous iteration. We refer to the quantity $\Gamma^t$ as the \textit{movement cost}.
In our image analysis pipeline experiment (Section \ref{sec:brain}), costs can be thought of as the amount of time  or the amount of compute required to re-run a specific module and all of the subsequent modules that follow. In this case, our goal is to perform an end-to-end optimization on the system to maximize the accuracy on a validation set, which can be measured with an f1-score or some other measure of the accuracy of the segmented output.

To trade-off between cost efficiency and functional optimality, we define the \textit{movement regret} as,
\begin{align}\label{moveregret}
    R^{+}(T,\lambda)=\sum_{t=1}^T f(\mathbf{x}^t)-f^* +\lambda\Gamma^t.
\end{align}
$\Gamma^t$ serves as a regularizer which is added to the standard definition of the cumulative regret.
In general, the function value and the cost are measured in different units, so $\lambda$ should depend on the scales of data.

\begin{figure*}[!t]
  \centering
    \includegraphics[width=0.92\textwidth]{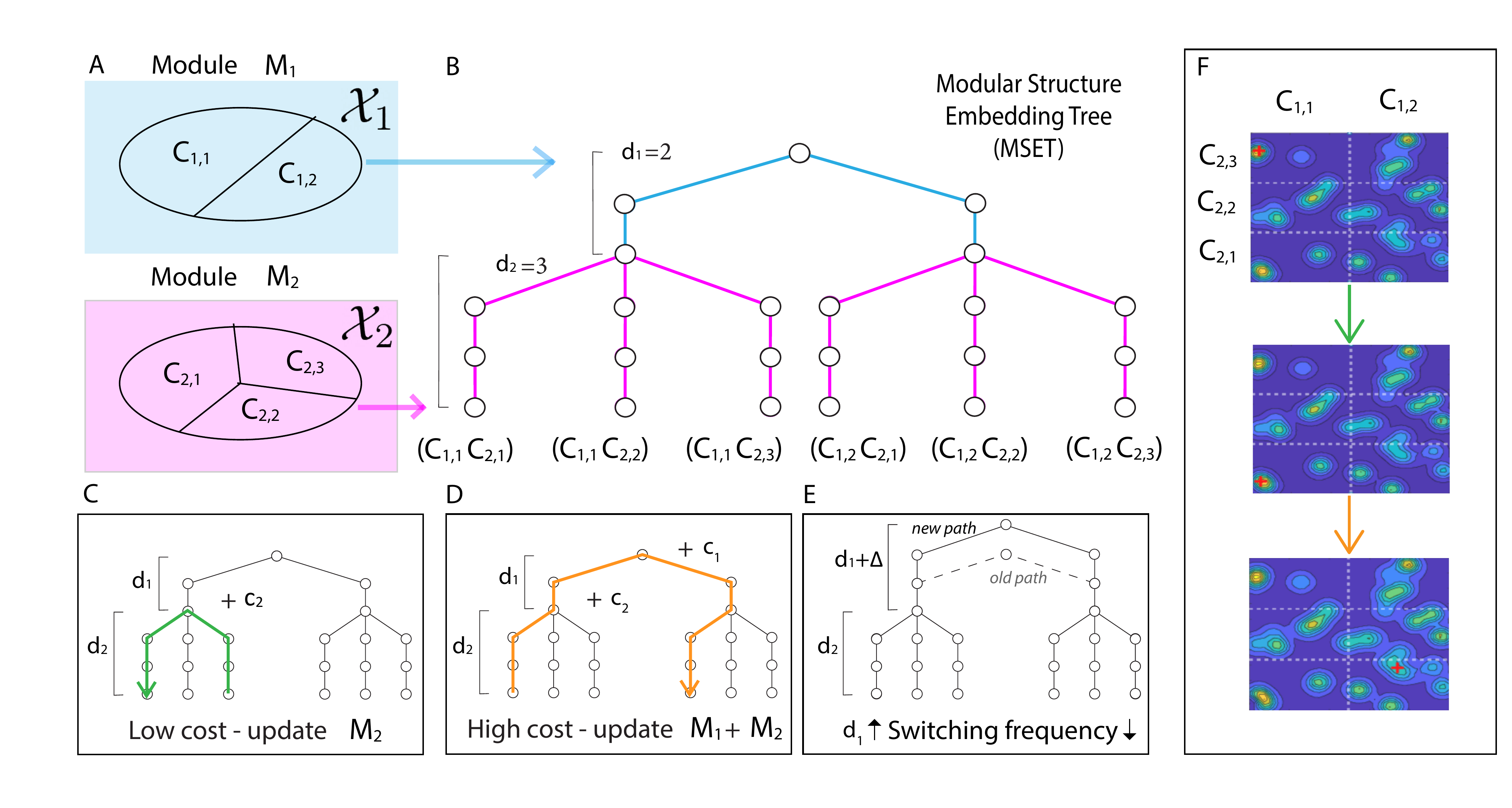}  
    \vspace{-1mm}
    \caption{\footnotesize {\em Overview of our approach}. 
     Illustration of the optimization in a modular system. In (A), we show a partition of variable spaces into regions and its corresponding MSET (B), constructed based on the partition and modular costs. An illustration of how changing regions incurs different costs (C-D), where in each case we trace the path between different arms. Changing the depth parameter $d_2\leftarrow d_2+\Delta$ produces a longer distance between any two arms and gives less incentive for arm changes (E). In (F), the landscapes of the BO update within regions at three consecutive iterations, corresponding to the arm changes in (C) and (D), respectively}
    \label{fig:mobo}
        \vspace{-3mm}
\end{figure*}

\vspace{-1mm}

\subsection{Algorithm}\label{sec:algorithm}
This section we provide the descriptions of the $3$ steps in the proposed algorithm~\ref{alg:main} named Lazy Modular Bayesian Optimization (LaMBO). 
%In Algorithm~\ref{alg:main}, we provide pseudocode for 
%\alg. This algorithm uses a SMB strategy \citep{pmlr-v65-koren17a} to impose switching costs on the sampling system and couples this method with a BO strategy to achieve our objective. 

%In this example, we depict a system that consists of two modules, partitioned into two and three sub-regions (subsets), respectively ({Figure~\ref{fig:mobo}B}). This partitioning of the space is then translated into a tree ({Figure~\ref{fig:mobo}C}), which encodes the cost to switch variables based upon the distance between the two partitions ({Figure~\ref{fig:mobo}D-F}), each represented as nodes on the tree. Finally, after selecting a joint variable subset of the space (arm) we use a BO strategy to estimate the underlying function of interest using Gaussian Process regression within each leaf (visualized in {Figure~\ref{fig:mobo}G} on the right). We now step through the details of the proposed approach (see Algorithm~\ref{alg:main}).

\paragraph{Step 1) Modular Structure Embedding Phase.} To visualize our algorithmic approach, we point the reader to {{Figure~\ref{fig:mobo}}}. 
In the first stage of our optimization procedure, we need to encode the switching costs associated with the system of interest. To do this, we take inspiration from the SMB algorithm described in Section~\ref{subsec:smb1} 
to encode the cost to switch variables using a tree-based approach (Figure~\ref{fig:mobo}B). We start by linking each arm with a region (subset) of variable space. The regions are flexible and can be partitioned in different ways, but should reflect the modular structure in the system. Thus, we choose to partition the variable space of each module separately. Specifically, ${\cal{P}}_m = \{{\cal{C}}_{m_1},{\cal{C}}_{m_2},\dots\ {\cal{C}}_{m_l}\}$ 
defines a partition for the $m^{\rm th}$ module, where ${\cal{X}}_{m} = \cup_{n} {\cal{C}}_{m_n}$.  
We require these sets to be disjoint ${\cal{C}}_{m n_1}\cap {\cal{C}}_{mn_2} = \emptyset$ for $n_1 \neq n_2$. 
Thus, when selecting an arm, we select a joint region of the first $N-1$ modules\footnote{We exclude the last module from partitioning procedure since the cost of changing parameters in the last module is the minimum cost per iteration, and can be changed freely at each iteration.}, i.e., $i \equiv({\cal{Z}}_1,\dots,{\cal{Z}}_{N-1})\in \mathcal{K}:={\cal{P}}_1\times\cdots\times{\cal{P}}_{N-1}$. %The loss associated with the $i^{\rm th}$ arm is then given by $\ell(i):=\min_{\mathbf{x}\in {\cal{Z}}_1,\dots,{\cal{Z}}_{N-1}\times {\cal{X}}_
%N}f(\mathbf{x})$. 

Next, we represent the arms in a tree $\mathcal{T}$ to encode the cost of switching between any two variable subsets. Intuitively, we want to build a tree that encodes the cost of switching between any two sets of hyperparameters (arms) in terms of the shortest path between these two leaves in the tree. Specifically, in Line $2$ of Algorithm~\ref{alg:main}, we call a subroutine ConstructMSET which returns a tree $\mathcal{T}$ (modular structure embedding tree, MSET), given a partitioning of the variables across all modules and depth parameters $d_m$, where $d_m$ is the depth of the m$^{\rm th}$ module. The partition and modular specification define the leaves of the tree and the depth parameters control the probability of switching, with higher depth in a module corresponding to lower switching probability (more laziness). In our example ( Figure~\ref{fig:mobo}B) , the tree consists of two parts (colored with blue and red) divided by the first forks, the upper portion corresponds to the partition of the first module, while the lower portion corresponds to the partition of the second module. In this case, the depth in the second module is set to 3 to reflect higher relative costs between the two modules and encourage lazy switching behavior.

\paragraph{Step 2) Optimization Phase.} Now the remaining task is to devise a strategy for arm selection and estimate the local optimum within its corresponding variable subset. We propose to use SMB for region (arm) selection, and then use a BO strategy to search within the selected region (Line $5-6$). The parameters of SMB and BO are updated at each iteration (Line $7-11$). Unfortunately, direct application of BO  changes all variables across each iteration, which typically incurs maximum cost. Hence, we propose an alternative \textit{lazy} strategy:
when the same variable subset is selected in an early module, we will use the results from the previous iteration rather than updating the outputs from this lazy module. This means that we do not need to rerun the module and thus can minimize the overall cost. Specifically, let $i_t$ be the arm we've selected and $({\cal{Z}}_1,\dots,{\cal{Z}}_{N-1})$ be its associated variable region. We propose to search for a block-wise update $\mathbf{x}^t= [\mathbf{x}_{1:m-1}^{t-1},\mathbf{u}]$ that minimizes the loss as follows:
\begin{align}\label{xt}
%\alpha_t(\mathbf{x}^t) =
\bar{\ell}_{t,0}(i_t): =\min_{\mathbf{u}\in {\mathcal{U}}} \alpha_t([\mathbf{x}_{1:m-1}^{t-1},\mathbf{u}]),
~~\mathcal{U} = (\prod\limits_{l=m}^{N-1}{\cal{Z}}_{l})\times {\cal{X}}_N,
\end{align}
where  $m$ is the first module that has a variable region that differs from the previous iteration $m:= \min \{n:{\cal{Z}}_{n}\neq{\cal{Z}}^{t-1}_{n}\}$, %$\bar{\ell}_{t,0}(i)$ defines the estimated loss associated with selecting arm $i$, 
and  $\alpha_t(\cdot)$ is a BO acquisition function.

\subsection{Empirical construction of MSET}
A crucial part of algorithm is the design of the subroutine \url{ConstructMSET}, which involves partitioning the variables in each module, and setting the depth parameters (${d_i}$'s). From our experiments, we observe that simple bisection aligned with coordinates yields good partition on many synthetic data and on our neural data. For a MSET with $|\cal{K}|$ leaves with the partition, \alg ~requires solving $|\cal{K}|$ local BO optimization problems per iteration. Hence initially, we partition each variable space of module to two subsets only, and abandon subsets when their arm selection probability $p_t$ is below some threshold after $10$ consecutive iterations. In our experiments, we always set the threshold to be ${0.1}/{|\cal{K}|}$, where $|\cal{K}|$ denotes the number of leaves of MSET. After that, we further divide the remaining  subsets again to increase the resolution. This procedure could be iterated upon further although we typically do not go beyond two stages of refinement. To avoid trapping in the local optimum, we also refresh the arm-selection probability and update the kernel hyperparameters simultaneously every $25$ iterations.

In our implementation, we set the depth parameter to be $d_i = 1$ or $d_i \propto \log {\lambda}c_i$ when $c_i$ could be estimated in prior. Empirically, we found that the performance is quite robust when $d_i\leq 5$ for the different cost ratios in both synthetic and real experiments we tested. To avoid accumulating cost too fast in early stages of \alg, we record the number of times that variable changes in the first module and 
dynamically increase the first depth parameters $d_1$ by $1$ every $20$ iterations when the the number has increased beyond $5$ ($1/4$ of the cycle) during the period. In all experiments, we have found this simple add-on perform on par or better than fixing depth parameters through an entire run. 
%%%%%%%%%% SHOULD WE INCLUDE?
%In practice, we found that the following improvements are useful.
%\noindent{\em 1. Restart with Epochs:} We refresh the arm-selection probability every $\tau$ iterations to escape from local optimum. In our implementation we choose $\tau = 25$ as the optional default value. 
%\noindent{\em 2. Adaptive Resolution Increase:} In this extension we discard the arms that have probability of selection being less than a threshold ($\tau=0.9$ in our implementation), and partition each remaining subset into $2$ subsets. We found that combining this with restart can accelerate the optimization in cases with many local optima. 

%\noindent{\em 3. More sophisticated partitioning methods:} 
%To incorporate domain knowledge is to put a prior scores on the variable partitions. 
%For instance, in our brain mapping example, there are certain combinations of parameters that would violate  size constraints related to the underlying biology. These unlikely parameters or parameter configurations could be incorporated into the design of the MSET. In this case, we added a regularized constant to the loss functions proportional to the averaged values observed in the partition in the similar data. bias the initial decisions.  The tree-based method \textit{High Confidence Tree (HCT)} \citep{azar2014online} is also a promising direction to rate partitions on the confidence of optimum occurence. }
%\eva{this last part is kinda weak, anything more interesting?}

\begin{algorithm}[t]
\begin{algorithmic}[1]
\caption{Lazy Modular Bayesian Optimization }\label{mobo}\label{alg:main}

\STATE {Input: }$\eta$, {\color{black}GP($\mu_0,k_0$), Partitions $\{\mathcal{P}_m\}_{m=1}^{N-1}$, depth parameters $\{d_m\}_{m=1}^{N-1}$}. 
\vspace{0.5mm}
\STATE{\color{black} $\mathcal{T}${ = \url{ConstructMSET}}($\{\mathcal{P}_m\}_{m=1}^{N-1}$,$\{d_m\}_{m=1}^{N-1}$).}
\vspace{0.5mm}
\STATE $H =$ depth($\cal{T}$), $\mathcal{K}=$ set of leaves, $p_1=\text{Unif}(\mathcal{K})$, $h_0=H$ and $i_0\sim p_1$.
\vspace{0.5mm}
 \FOR {$t=1$ to $T$}
\STATE Select arm $i_t\sim p_t(\cdot|A_{h_{t-1}}(i_{t-1}))$. 
\vspace{0.5mm}
\STATE {\color{black}Choose $\mathbf{x}^t$ by solving Eq.~(\ref{xt}).} 
\vspace{0.5mm}
\STATE Let $\sigma_{t,h}$, $h=1,\dots,H-1$, be i.i.d. Unif($\{-1,1\}$). 
\vspace{0.5mm}
\STATE let $h_t=\min\{0\leq h\leq H:\sigma_{t,h}=-1\}$ where $\sigma_{t,H}=-1$. 
\STATE Obtain loss estimators via $\tilde{\ell}_t~=$~Eq. (\ref{recursiveloss}), Eq.~(\ref{xt}) and
\STATE $$p_{t+1}=\frac{p_t(i)e^{-\eta\tilde{{\ell}}_t(i)}}{\sum_{j=1}^{|\mathcal{K}|}p_t(j)e^{-\eta\tilde{{\ell}}_t(j)}},~\forall i\in \mathcal{K}.$$ %\COMMENT{Update Weight}
\vspace{1mm}
\STATE {\color{black}Posterior Updates by Eq.~(\ref{postupdate}).}
\ENDFOR
\end{algorithmic}
\end{algorithm}

\section{Algorithmic Analysis}

\label{sec:analysis}

In this section, we analyze the performance of Algorithm \ref{alg:main} from two perspectives: {\em 1. Optimization accuracy} and {\em2. Cost efficiency}. Our main result, which is stated in Theorem \ref{mainthm}, shows that \alg~achieves sublinear movement regret when the parameters of the input tree are set properly using the cost structure of the system. 

Our results are presented in terms of \textit{maximum information gain} defined below.

\textbf{Definition 1.} \textit{\textbf{Maximum Information Gain.}~
Let $f\sim GP$ be defined in the domain $\cal{X}$. The observation of $f$ at any $\mathbf{x}$ is given by the model $y=f(\mathbf{x}) +\epsilon$, $\epsilon \sim {\cal{N}}(0,\sigma)$. For any set $A\in \cal{X}$, let $f_A$ and $y_A$ denote the set of function values and observations at points in $A$, and $I$ denote the Shannon Mutual Information. The Maximum Information Gain is defined by 
$\gamma_T:=\max_{A\subset {\mathcal{X}}:|A|=T} I(y_A,f_A)$
}

Analytical bounds on $\gamma_T$ of common kernels are provided in Supp. \ref{subsec:boundmui}. 
To proceed with our analysis, we make the following assumption on the objective function.

{\textbf{Assumption 1.}}~{\em The function $f$ is $L$-Lipschitz, non-negative, and has a bounded norm $\|f\|_{{\mathcal{H}}_{k_0}}\leq 1$ in the reproducing kernel Hilbert space  ${\mathcal{H}}_{k_0}$.}

Note that our assumption is not too stringent since for any function in a Hilbert space defined above, 
$L$ can be estimated by $|f(x)-f(y)|\leq \|f\|_{\cal{H}}\|\Phi(x)-\Phi(y)\|_{\cal{H}}$; for instance, $L=1/w$ for exponential kernel $k_0(\mathbf{x},\mathbf{x}') = \exp(-\|\mathbf{x}-\mathbf{x}'\|^2/w^2)$ since $\|\Phi(x)-\Phi(y)\|_{\cal{H}}\leq \|\mathbf{x}-\mathbf{y}\|/w$.

\subsection{Optimization Accuracy}

%The most untypical assumption might be the non-negativity of the objective, but it holds under various application, e.g. validation accuracy or f1-score.

Our first lemma concerns the optimization capability of \alg~under a common  definition of regret, $R(T):=
\sum f(x_t) -f^*$. Note that we choose to represent it in expectation instead of a probability bound for notational compactness. Conversion to one another is straightforward by common technique like Markov inequality.
\begin{lemma}\label{lem:cumuregret} \textbf{Ordinary Regret Bound.}
Suppose the learning rate of the \alg~is set to be $\eta=\sqrt{ 2^{-H}T^{-1}\log |\mathcal{K}|}$, where $H$ is the depth of the MSET, then the expected cumulative regret of  \alg~is: 
\begin{align*}
     \mathbb{E}[R(T)]={\cal{O}}\left(\sqrt{2^HT\log |\mathcal{K}|}\right).
\end{align*}
\end{lemma}
\begin{remark}
By treating modules as arms, a natural comparison is the result of SMB in \citep{pmlr-v65-koren17a} where $\mathbb{E}[R_T]={\cal{O}}\left(\sqrt{kT\log |\mathcal{K}|}\right)$.
As the arms $|\mathcal{K}|$ are represented as leaves of a binary tree with depth $H$, we must have $2^H\leq k$, which shows that the regret bound we have is upper bounded by the result of SMB. The equality holds when the tree is complete. On the other hand, the gap between us could be potentially large. The key to this improvement is by leveraging arm correlation; by using Gaussian surrogate, each sample gives the information of not only the pulled arm itself, but also that of infinitely many others. 
\end{remark}

\subsection{Analysis of Cost Efficiency}
Next, we analyze the cost incurred by adopting \alg.
The following lemma shows that \alg~is capable of accumulating sublinear cost. The result also gives an explicit recipe of choosing parameters $\{d_i\}$ of MSET from theory.
Below we provide a sketch of the achievable rate and defer the detailed forms of parameters to Supp. \ref{sec:suppproof}.
\begin{lemma}\label{lem:cost} \textbf{Cumulative Switching Cost.}
For sufficiently large $T$, there exists depth parameters 
 $\{d_i\}$ of the MSET such that 
 \alg~accumulates movement cost
\begin{align*}
\mathbb{E}[\sum_{t=1}^T\Gamma^t]
    = \mathcal{O} \bigg( \sum\limits_{m=1}^{N-1}c_mT^{2/3}\log |\mathcal{K}|\log \frac{T^{1/3}}{\log|\mathcal{K}|} \bigg). 
\end{align*}
\end{lemma}
\begin{remark}
A striking implication of Lemma 2 is that even with nonconstant cost $c_m = \Omega(1)$, the cumulative cost could still be sublinear as long as $c_m \ll o(T^{1/3})$.
\end{remark}
Finally combining the above two lemmas leads to our concluding theorem, which shows that a simple partition strategy, along with proper selection of the depth parameters $d_i$, gives sublinear movement regret defined in (\ref{moveregret}). 
Without additional information about how to partition each module, the simplest way to partition the space is uniformly. Hence in the analysis we adopt an uniform partition strategy characterized by ${r_i}$, where $r_i$ denotes the Euclidean diameter of the partitioned subset $\mathcal{X}_i$.

Now we present a sketch of our main theoretical result
where a proof and detailed constants could be found in Supp. \ref{sec:suppproof}.
\begin{theorem}\label{mainthm} \textbf{Movement Regret Bound.}
For $1\leq m\leq N-1$, let $D_m$ denote the dimension of ${\cal{X}}_m$. Suppose for all $t>0$, $1\leq m \leq N-1$, we set $\beta_t=\Theta(\sqrt{\gamma_{t-1}+\ln T})$,  $\eta=\Theta(T^{-2/3}\sum_{m=1}^{N-1}D_m\log(LT^{1/3}/{D_m\log T}))$. The MSET has uniform partition of each ${\mathcal{X}}_m$ with diameters
$r_m=\frac{D_m}{L}T^{-\frac{1}{3}}\log T$, where the depth parameters $d_m$ are chosen according to Lemma \ref{lem:cost}, and UCB acquisition function is used.
Then \alg~achieves the expected movement regret
\begin{align*}
   \mathbb{E}[ R^+]=\mathcal{O}((\lambda\sum\limits_{j=1}^{N-1} c_j\sum\limits_{m=1}^{N-1}D_m T^{\frac{2}{3}}(\log T)^2) +\gamma_T\sqrt{T}).
\end{align*}
\end{theorem}
\begin{remark}
{\bf Comparisons to moving bandit algorithm:} A black-box optimization strategy blind to switch cost usually has $\mathbb{E}[R(T)] = o(1)$, $\mathbb{E}[\sum_{t=1}^T\Gamma^t] = \Omega(T)$ and thus obtain a linear movement regret $\mathbb{E}[R^+]=\mathbb{E}[R(T)]+\mathbb{E}[\sum_{t=1}^T\Gamma^t] =\Omega(T)$.
For switch-cost-aware alternatives, the closest result on moving regret is Theorem $2$ in \citep{pmlr-v65-koren17a}. However, their result relies on the Lispchitz property of the movement metric, which does not hold in our setting as the cost from changing variables in modules is not even continuous. By leveraging arms correlation with BO and adapting a lazy arm selection strategy, we extend their result by achieving a sublinear rate. 
\end{remark}
%\begin{remark}
%{\bf Dominant effect on the regret: }In Theorem $1$, the two terms contributing to the regret arise from bandit and Bayesian optimization, respectively. It immediately follows from the bounds above that the regret is dominated by the former for linear and squared exponential, and Mat\'ern (if $D(D+1)<4\nu$) kernels.
%\end{remark}

\begin{figure*}[t!]
\centering
   \includegraphics[width=\textwidth]
   {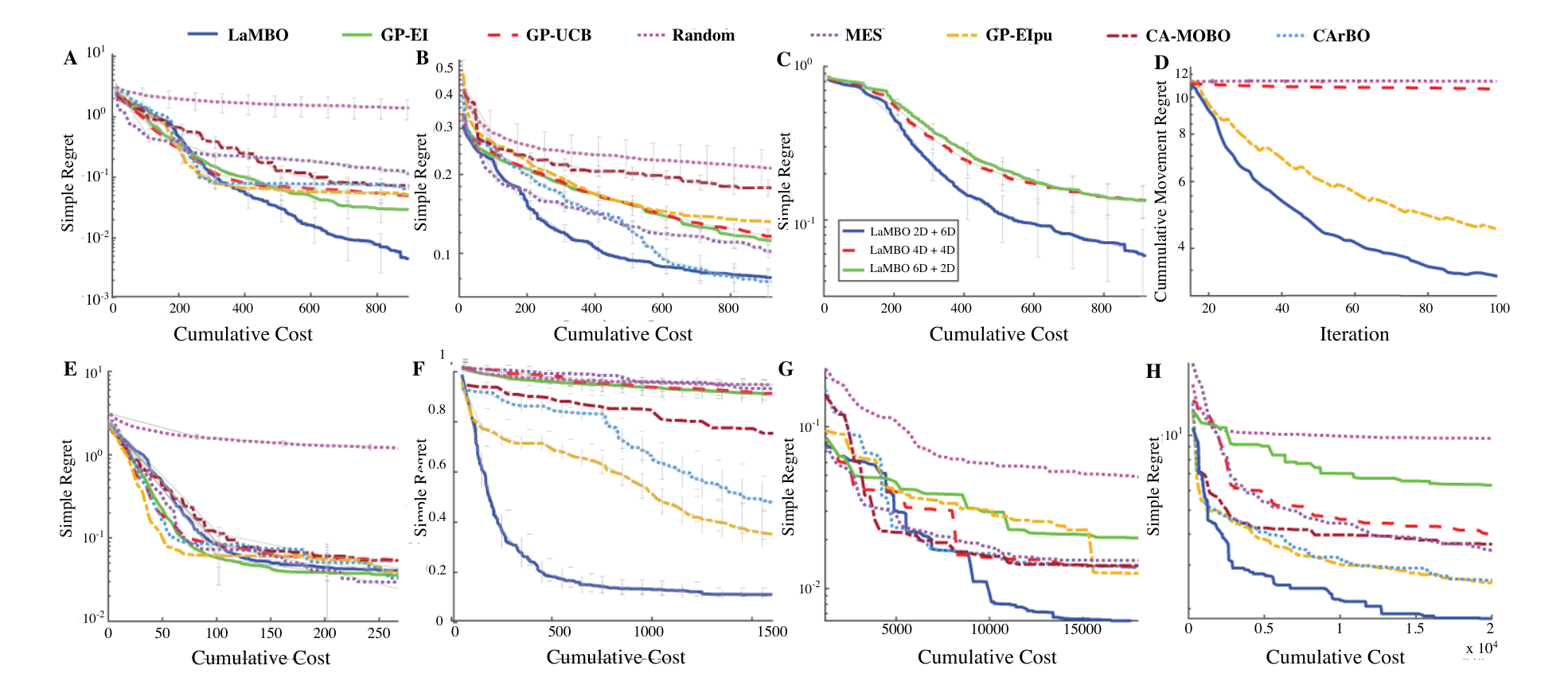}
   \vspace{-1mm}
\caption{\footnotesize {\em Results on synthetic datasets and a brain mapping example.} 
In (A-D), we compare 
\alg~with other BO algorithms on synthetic functions in a two module setting with a cost ratio of $10$ to 1. In (A-B), we show the results for two different synthetic functions (Hartmann 6D, Rastrigin 6D). We split the variables of them into two modules with the first 3 dimensions in one module and the remaining three in the second. 
%%We also split the 6D Griewank  function into a module with 4 dimensions and then 2D in the second, and finally split the 8D Ackley function into 6 and 2 dimensions.
In (C), we study the impact of splitting variables into sets of different dimensions, by splitting the Ackley 8D into three different configurations $[2,6]$, $[4,4]$, and $[6,2]$.
In (D), we study the cumulative movement regret with Hartman 6D. This verifies our theory that \alg~can effectively reduce the movement regret. In (E), we show the performance of the different approaches when the cost ratio between modules equals one, using Hartman 6D. (F) explores the three module setting of Ackley 8D splitting the variables into dimensions $[2,2,4]$, and define the costs by $=[40,10,1]$. In this case, when the costs accumulate early on, \alg~  really shines. Finally, we depict a brain mapping pipeline consisting of two (G) and (H) three modules, where the costs are modeled with estimated amount of time to execute each module ([326,325,55] sec).}
\label{fig:syn}
    \vspace{-3mm}
\end{figure*}

\vspace{-6mm}
\section{Experiments}\label{sec:exp}
In this section, we start by testing \alg~on benchmark synthetic functions used in other studies \citep{vellanki2017process,kirschner2019adaptive}. Following this, we apply \alg~to tune a multi-stage neuroimaging pipeline that reconstructs 3D images from segmented 2D images.
\paragraph{Experimental setup.}
For simplicity, we used the squared exponential kernel and initialized it using $15$ random samples before starting the inference procedure.
In our experiments, the functions are normalized by their maximized absolute value for clear comparisons,  the regularization parameter is fixed to $\lambda=0.1$, the UCB parameter is set each iteration as $\beta_t=0.2D\log 2t$, and the learning rate is set to $\eta = 1$. The sampling noise $\epsilon$ is assumed to be independent Gaussian with standard deviation $0.01$. For construction of MSET, we test on the simplest case where $d_i=1$ and partition the variable space in each module into 2 sets aligned with a random coordinate. Some practical and detailed discussions on the hyperparameter choices and partition strategies are deferred to Supp.
\ref{sec:practical}.
The curves on synthetic data and real data were computed by averaging across $100$ and $20$ simulations, respectively. We compare \alg~ with common baselines GP-UCB \citep{Srinivas:2010:GPO:3104322.3104451}, GP-EI \citep{movckus1975bayesian}, Max-value entropy search \citep{wang2017max},  random sampling, and three cost-aware strategies: EIpu \citep{snoek2012practical}, CA-MOBO \citep{abdolshah2019cost}, and CArBO \citep{lee2020costaware}. 
To adapt the cost-aware strategies to our setting, we  update the cost function at each iteration to be to the switching cost 
$\Gamma^t$ defined in Eqn.~(\ref{eq:movementcost}) in Section \ref{sec:Approach}.

\subsection{Experiments on synthetic functions}
\vspace{-2mm}
For synthetic benchmarks, we selected a number of common functions used to test algorithms in the literature. However, unlike our real data examples that have clear modular structure due to the different sets of operations performed at different stages, the variables in synthetic test functions do not readily admit a modular structure. Thus  to simulate a $2$-module or $3$-module scenario, we divide the variables in each function into different groups to create effective modules. 
%To emphasize the the effect of accumulation of cost in early stages of processing, we set the cost ratio between the first and second module to be 10 to 1.

In Figure~\ref{fig:syn} (A-D), we compare 
the methods on synthetic functions in a two-module setting with a cost ratio of $10$ to $1$. In (A-B), we show the results for two different synthetic functions Hartmann and Rastrigin, respectively (more experiments on synthetic functions could be found in Supp. \ref{supp:sec:syn}). In (C), we study the impact of splitting variables into sets of different dimensions with function Ackley 8D, a synthetic function with a sharp global optimum surrounded with multiple local ones. 
The result suggests \alg~is stable among different variable configurations in modules.
Under the same setting as in (A), we verify our regret analysis by plotting the cumulative movement regret curve in (D), and study the performance of the different approaches when the cost is $[1,1]$. The former shows that \alg~minimizes the averaged movement regret of (\ref{moveregret}) better than cost-aware and unaware baselines. The later shows that \alg~performs better when the cost ratio between modules is large while on par with alternative when the ratio is $\simeq 1$. (F) explores the $3$-module setting of Ackley 8D ($[2,2,4]$), with the cost $[40,10,1]$. In this case, we found that it performs even better than its $2$-module counterpart in Figure \ ref{fig:synsupp}, suggesting \alg's applicability in pipeline with many modules. 

Overall, we find that \alg~outperforms other approaches and really shines when the cost of earlier modules is much larger (as seen in (A) vs. (E)). 
%We also observe sharp global optimum among multiple local ones. 
%1This could be explained by the stochastic nature of SMB mechanism which advocates for more exploration. 
When we track the optimization trajectory, we observe that \alg~performs similarly to other methods early on, but with further iterations, \alg~starts to outperform the alternatives. This could be explained by inaccurate estimation of the function at early stages, and the fact that aggressive input changes could outperform the more conservative or lazy strategy used in \alg. However, as more samples are gathered, \alg~demonstrates more power in terms of its cost efficiency by being lazy in variable switching. 
%We can obtain further insights by examining the cumulative movement regret (Figure~\ref{fig:syn}D) for our method vs. cost-aware baselines. Our results reveal that \alg~ has minimal regret and cost-aware methods have intermediate results between non-cost-aware strategies and ours. 

\subsection{Application to a multi-stage neuroimaging pipeline}

\label{sec:brain}
Segmentation and identification of neural structures of interest (e.g., cell bodies, axons, or vasculature) is an important problem in connectomics and other applications of brain mapping \citep{helmstaedter2013connectomic,oh2014mesoscale,dyer2017quantifying}. However, when dealing with large datasets, transfer can be challenging, and thus workflows must be re-optimized for each new dataset \citep{johnson2020toward}.  Here, we consider the optimization of a relatively common three-stage pipeline, consisting of a pre-processing (image enhancement via denoising and contrast normalization), semantic segmentation for pixel-level prediction (via an U-net architecture), and a post-processing operation (to reconstruct a 3D volume). For comparison, we also consider a simplified pipeline without the pre-processing.
To optimize this pipeline, we use a publicly available X-ray microCT dataset \citep{prasad2019mapping} to set up the experiments in both a two-module (no pre-processing) and full three-module version of the pipeline.

In the first module, a pre-processing operation is performed where we tune a contrast parameter and denoising parameter. 
In the second module we train an U-Net, where in this case we tune the learning rate and batch size. The third module is in charge of post-processing and generates 3D reconstructions from the U-Net output; the hyperparameters in this module include a label purity score, cell opening size, and a shape parameter to determine whether uncertain components are either cells or blood vessels.
Details of search space for each module are described in Supp. \ref{sec:brain}). The cost of the experiment is the aggregate recorded clock time for generating an output after changing a variable in a specific module. To test \alg~on the problem, we gathered an offline data set consisting of $606,000$ different hyperparameters obtained by exhaustive search.

%In our experiment, the best set of hyperparameters in this case achieved an f1-score of $77.129\%$, while the averaged random score is below $50\%$.
In the two-module case ({Figure~\ref{fig:syn}G}), we observe a transition effect; when enough cost has been spent, \alg~starts to increase its gap in performance over other methods. In the three-module case ({Figure~\ref{fig:syn}H}) the advantage is even more pronounced, where the transition happens earlier. 
Quantitatively it shows that to get close to the optimum (within 5\%), \alg~can achieve this result in only 25\% of the time required by the best alternative approach (1.4 vs. 5.6 hours).

%\caption{\label{fig:recons}\footnotesize {\em Results from the three stage pipeline for an optimal set of parameters and a suboptimal set.} From from left to right, we show a pre-processed input image example, a U-Net output for the same example, and a post-processed 3D reconstruction. 
%The f1-score measured in step 3 is $77.129\%$. 
%Along the top row (A), we show the results obtained for an optimal set of hyperparameters selected by our approach (as measured by the f1-score). Along the bottom row (B), we show the same results for a suboptimal hyperparameter combination with poor performance. 
%The mean and standard deviation across 100 random hyperparameters was computed for each module, $326.059 \pm 128.390$ sec (pre), $324.539 \pm 128.367$ sec (U-net), and $54.773 \pm 0.894$ sec (post).
%The f1-score measured in step 3 is only $27.692\%$.
%The hyperparameters used in each step are indicated along the top of each row of images, and the average f1-score achieved by that combination is indicated under each row. 

%\label{fig:image}
%\vspace{-6mm}
%\end{figure}

\section{Discussion}\label{sec:discussion}

%This paper addresses a real-world problem of system optimization that is encountered in a variety of scientific disciplines.

%In this paper, we introduced a new algorithm for Bayesian optimization that leverages known modular structure in an otherwise black-box system to minimize the overall cost required for global optimization. 
%We demonstrated the applicability of our algorithm both in theory and in practice. 

This paper addresses a real-world problem of system optimization that is encountered in a variety of scientific disciplines. Increasingly, as we expand the size of datasets in different domains, we need automated solutions to quickly apply advanced machine learning systems to new datasets and re-optimize  systems in an end-to-end manner. 
To tackle this problem, we introduced a new algorithm for Bayesian optimization that leverages known modular structure in an otherwise black-box system to minimize the overall cost required for global optimization. 
We showed how to leverage structure in such systems by incorporating a lazy switching strategy with Bayesian optimization. 
%We demonstrated its application in a relatively simple system. 
In the future, we would like to generalize our method to the case where both the function and switching costs are unknown, and extend to more complex cost hierarchies. 

\section*{Acknowledgements}
\vspace{-2mm}
This work was supported by the NIH award 1R24MH114799-01 and  awards IIS-1755871 and CCF-1740776 from the NSF.

%\section*{Code availability}
%The code is provided here: \url{https://anonymous.4open.science/r/9328c3c5-35ba-46a2-aa79-031d1de6836d}. \eva{@henry - we can make it available on github? we dont need to have double blind, right??}

\bibliography{lin_383-cbo.bib}

\onecolumn
\section*{\large Supplementary Materials}

\renewcommand{\thefigure}{S\arabic{figure}}
\setcounter{figure}{0}

\appendix

\section{Technical Preliminaries and Proofs}\label{sec:suppa}

\subsection{Proofs of Theoretical Results}\label{sec:suppproof}
We begin with the preliminary Lemmas \ref{lem:beta} and \ref{lem:sigsum}. 
\begin{lemma}\label{lem:beta}(Theorem $2$ of \citep{Chowdhury:2017:KMB:3305381.3305469}). Let $f$ be a function lying in the RKHS ${\cal{H}}_{k_0}$ of kernel $k_0$ such that $\|f\|_{{\cal{H}}_k}\leq 1$ with input dimension $D$. Assume the process of observation noise $\{\epsilon_t\}$ is $\sigma$-sub-gaussian. Then setting
\begin{align*}
    \beta_t =   1+\sigma \sqrt{2\left(\gamma_{t-1}+1+\ln (1 / \delta)\right)},
\end{align*}
we have the following holds with probability at least $1-\delta$,
\begin{align*}
    |\mu_{t}(x)-f(x)|\leq \beta_{t+1}\sigma_{t}(x), \quad \forall x\in\mathcal{X}, \quad\forall  t\geq 1,
\end{align*}
where $\mu_{t},\sigma_t$ are given by the formula
\begin{align*}
&\mu_{t}(\mathbf{x})=\mathbf{k}_{t-1}^T(\mathbf{x})(\mathbf{K}_{t-1}+\sigma \mathbf{I})^{-1}\mathbf{y}_{t-1},\\
&\sigma_{t}^2(\mathbf{x})=k_0(\mathbf{x},\mathbf{x})-\mathbf{k}_{t-1}^T(\mathbf{x})(\mathbf{K}_{t-1}+\sigma \mathbf{I})^{-1}\mathbf{k}_{t-1}(\mathbf{x}).
\end{align*}
\end{lemma}

\begin{lemma}\label{lem:sigsum}(Lemma 4 of \citep{Chowdhury:2017:KMB:3305381.3305469}). Suppose we sample the objective function $f$ at $\{\mathbf{x}^1,\dots,\mathbf{x}^{T-1}\}$ then the sum of standard deviations is bounded by,
\begin{align*}
    \sum_{t=1}^{T} \sigma_{t-1}\left(\mathbf{x}^{t}\right) \leq \sqrt{4(T+2) \gamma_{T}}.
\end{align*}
\end{lemma}

Our first goal is to prove Lemma \ref{lem:cumuregret}.
\begin{replemma}{lem:cumuregret}
Suppose the learning rate of the \alg~is set to be $\eta=\sqrt{ 2^{-H}T^{-1}\log |\mathcal{K}|}$, where $H$ is the depth of the MSET, then the expected cumulative regret of  \alg~is: 
\begin{align*}
     \mathbb{E}[R(T)]={\cal{O}}\left(\sqrt{2^HT\log |\mathcal{K}|}\right).
\end{align*}
\end{replemma}
\begin{remark}
We can compare the result with SMB in \citep{pmlr-v65-koren17a} where $\mathbb{E}[R_T]={\cal{O}}\left(\sqrt{kT\log |\mathcal{K}|}\right)$. Note that ours is a lower bound of it as $2^H\leq k$. They could potentially have a large gap between them in terms of order. This performance improvement is due to our loss estimator adapted to arm correlation, whereas
\citep{pmlr-v65-koren17a} considers the pure bandit information.
\end{remark}

The following Lemma \ref{smblem} is a key to prove Lemma \ref{lem:cumuregret} in the main text.
\begin{lemma}\label{smblem}
For any sequence of ${\tilde{\ell}}_{1},$ $\dots,$ ${\tilde{\ell}}_{T}$, denote $i^*$ to be the solution of $\max_i\sum_{t=1}^T {\tilde{\ell}}_{t}(i)$ and assume $2^H\leq c\frac{T}{\log |K|}$ for some constant $c>0$, then there exists an $\eta=\Theta(\sqrt{2^{-H}T^{-1}\log |\mathcal{K}|})$ such that \alg~has the property
\begin{align*}
    \mathbb{E}[\sum_{t=1}^T (p_t \cdot {\tilde{\ell}}_{t} - {\tilde{\ell}}_{t}(i^*))] \leq \frac{\log |K|}{\eta} + \eta T 2^{H+1}.
\end{align*}
\end{lemma}

For the proof of \ref{smblem} we follow the path in \citep{NIPS2017_7000}. Before we start the proof of Lemma \ref{smblem}, we will need Lemma \ref{lossbound}, \ref{bias}, \ref{variance}, and \ref{regretb}.

\begin{lemma}\label{lossbound}
\begin{align}\label{fineq}
0\leq \bar{\ell}_{t,h}(i)\leq \prod_{j=0}^{h-1}(1+\sigma_{t,j}), \quad\forall i \in K.
\end{align}
In particular, if $\sigma_{t,h}=-1$ then $\bar{\ell}_{t,j}=0$ for all $j>h$.
\end{lemma}
\begin{proof}
The last statement is trivial from definition.
We will prove Eq.~(\ref{lossbound}) by induction on $h$. Since $0\leq \bar{\ell}_{t,0}(i)\leq 1$ by Eq.~(\ref{xt}) and the UCB is upper bounded by $1$, the statement holds for $h=0$. Now assume it holds for $h-1$. Then firstly,
\begin{align*}
    \bar{\ell}_{t,h}(i)\geq -\frac{1}{\eta}\log\left(\sum_{j\in A_h(i)}\frac{p_t(j)}{p_t(A_h(i))}\right)=0.
\end{align*}
Secondly, applying Jensen's inequality, we have
\begin{align*}
    \bar{\ell}_{t,h}(i)&\leq -\frac{1}{\eta}\sum_{k \in A_h(i)}\frac{p_t(k)}{p_t(A_h(i))}\log (\exp(-\eta(1+\sigma_{t,h-1}){\bar{\ell}}_{t,h-1}(k)))\\
    &= (1+\sigma_{t,h-1})\sum_{k \in A_h(i)}\frac{p_t(k)}{p_t(A_h(i))}\bar{\ell}_{t,h-1}(k)\nonumber\\
    &\leq\sum_{k \in A_h(i)}\frac{p_t(k)}{p_t(A_h(i))}\prod_{j=0}^{h-1}
    (1+\sigma_{t,j})\nonumber\\
    &=\prod_{j=0}^{h-1}(1+\sigma_{t,j}),
\end{align*}
where the second inequality is followed by the induction assumption. Therefore the proof is complete by mathematical induction.
\end{proof}

\begin{lemma}\label{bias}
For all $t$ and $0\leq h\leq H$ the followings hold:
\begin{itemize}
\item For all $t$ we have
\begin{align}
    \mathbb{E}\left[p_t\cdot \tilde{\ell}_{t}\right]=\mathbb{E}\left[
    \tilde{\ell}_{t}(i_t)\right].
\end{align}
\item With probability at least $1-2^{-(h+1)}$, we have that $A_h(i_t)=A_h(i_{t-1})$.
\end{itemize}
\end{lemma}
\begin{proof}
The proof of the second property is identical to Lemma 8 in \citep{NIPS2017_7000} and is thus omitted . Now we prove the first property. Note that we only need to prove 
\begin{align}
    \mathbb{E}[\textbf{1}(i_t=i)]=\mathbb{E}[p_t(i)], \quad \forall t>0,\quad \forall i \in \mathcal{K}.
\end{align}
We will again use the mathematical induction to prove the above statement.
The initial case $t=1$ holds trivially. Now assume the statement is true for $t=k$. Then for $t=k+1$,
\begin{align*}
    &\mathbb{E}[\textbf{1}(i_{k+1}=i)|h_k=0]=\mathbb{E}[\textbf{1}(i_{k}=i)|h_k=0]\\
    =&\mathbb{E}[\textbf{1}(i_k=i)]=\mathbb{E}[p_k(i)],
\end{align*}
where the last equality follows from the induction assumption.
On the other hand,
\begin{align*}
    \mathbb{E}[p_{k+1}(i)|h_k=0]=\mathbb{E}[p_k(i)|h_k=0]=\mathbb{E}[p_k(i)],
\end{align*}
where the last equality follows from the independence between $p_k$ and $h_k$.
Hence we have 
\begin{align}\label{h=0}
    \mathbb{E}[\textbf{1}(i_{k+1}=i)|h_k=0]=\mathbb{E}[p_{k+1}(i)|h_k=0].
\end{align}
Now if $h_k=h'>0$. Let $A'\in{\cal{A}}_{h'}$ be the subtree such that $\{i\}\subset A'$ then by the tower rule for expectation we have
\begin{align*}
    &\mathbb{E}[\textbf{1}(i_{k+1}=i)|h_k=h',p_{k+1}]\\
    =&\mathbb{E}[\textbf{1}_{A'}(i_{k+1})\mathbb{E}[\textbf{1}(i_{k+1}=i)|h_k,p_{k+1},i_k\in A']|h_k=h',p_{k+1}]\nonumber\\
    =&\mathbb{E}[p_{k+1}(A')p_{k+1}(i|A')|h_k=h',p_{k+1}]\nonumber\\
=&\mathbb{E}[p_{k+1}(i)|h_k=h',p_{k+1}].
\end{align*}
Therefore,
\begin{align}\label{alld}
    \mathbb{E}[\textbf{1}(i_{k+1}=i|h_k=h')]=\mathbb{E}[p_{k+1}(i)|h_k=h'].
\end{align}
By (\ref{h=0}), (\ref{alld}) now holds for every possible value of $h_k$, so we must have
\begin{align*}
    \mathbb{E}[\textbf{1}(i_{k+1}=i)]=\mathbb{E}[p_{k+1}(i)],
\end{align*}
which completes the proof by induction.
\end{proof}

\begin{lemma}\label{variance}
For all $t$, we have $\mathbb{E}[p_t\cdot {\tilde{\ell}}_{t}^2]\leq 2^{H+1}$.
\end{lemma}
\begin{proof}
Observe
\begin{align*}
    {\tilde{\ell}}^2_{t}(i)\leq \left(\bar{\ell}_{t,0}(i)+\sum_{h=0}^{H-1}\sigma_{t,h}\bar{\ell}_{t,h}(i)\right)^2.
\end{align*}
Since $\mathbb{E}[\sigma_{t,h}]=0$ and $\mathbb{E}[\sigma_{t,h}\sigma_{t,h'}]=0$ for $h\neq h'$, we have 
\begin{align}\label{2prop}
    \mathbb{E}[{\tilde{\ell}}^2_t(i)]\leq 2\sum_{h=0}^{H-1} \mathbb{E}[\bar{\ell}^2_{t,h}(i)].
\end{align}
Now by Lemma \ref{lossbound} we have
\begin{align*}
    p_t\cdot\bar{\ell}_{t,h}^2\leq \sum_{i\in \mathcal{K}}p_t(i)\prod_{j=0}^{h-1}(1+\sigma_{t,h})^2.
\end{align*}
Then taking expectation on both sides leads to
\begin{align}\label{2h}
    \mathbb{E}[p_t\cdot\bar{\ell}_{t,h}^2]\leq \sum_{i\in K}p_t(i) 2^h=2^h.
\end{align}
Finally, combining Eq.~(\ref{2prop}) with Eq.~(\ref{2h}), we get
\begin{align*}
    \mathbb{E}[p_t\cdot{\tilde{\ell}}_{t}^2]\leq 2\sum_{h=0}^{H-1}\mathbb{E}[p_t\cdot\bar{\ell}_{t,h}^2]\leq 2^{H+1}.
\end{align*}
\end{proof}

\begin{lemma}\label{regretb}\citep{alon2015online}. Let $\eta>0$ and  $\mathbf{z}_1,\dots,\mathbf{z}_T\in \mathbb{R}^{|\mathcal{K}|}$ be real vectors such that $\mathbf{z}_t(i)\geq -\frac{1}{\eta}$ then a sequence of probability vectors $p_1,\dots,p_T$ defined by $p_1=(1/|\mathcal{K}|,\dots,1/|\mathcal{K}|)$ and for all $t>1$,
\begin{align*}
    p_t(i)=\frac{p_{t-1}(i)\exp(-\eta \mathbf{z}_t(i))}{\sum_{j\in \mathcal{K}}q_{t-1}(j)\exp(-\eta \mathbf{z}_t(j))},
\end{align*}
have the property that
\begin{align*}
    \sum_{t=1}^Tp_t\cdot \mathbf{z}_t \leq \sum_{t=1}^{T}\mathbf{z}_t(i^*) +\frac{\log |\mathcal{K}|}{\eta} +\eta\sum_{t=1}^T p_t\cdot \mathbf{z}_t^2,
\end{align*}
for any $i^*\in \mathcal{K}$.
\end{lemma}

Now we are ready to prove Lemma \ref{smblem}.

\begin{proof}
By the assumption that $2^H\leq c\frac{T}{\log |K|}$ for some constant $c>0$, if we set $\eta=\sqrt{c^{-1}2^{-H}T^{-1}\log |K|}$ then we have $2^H\leq \frac{1}{\eta}$. Also observe that $\tilde{\ell}_t = \bar{\ell}_{t,0} +\sum_{j=0}^{h_t-1}\bar{\ell}_{t,j}-\bar{\ell}_{t,h_t}$, so Lemma \ref{lossbound} implies that $\bar{\ell}_{t}\geq -\frac{1}{\eta}$. Now we apply Lemma \ref{regretb} to the sequence $\{\tilde{\ell}_t\}_t$ to obtain
\begin{align}\label{51}
    \sum_{t=1}^Tp_t\cdot {\tilde{\ell}}_t-\sum_{t=1}^T{\tilde{\ell}}_t(i^*)\leq \frac{\log |\mathcal{K}|}{\eta} +\eta\sum_{t=1}^Tp_t\cdot {\tilde{\ell}}^2_t.
\end{align}
Finally, we take expectation on both sides of Eq.~(\ref{51}) together with Lemma \ref{bias} and \ref{variance}, then
\begin{align}
&\mathbb{E}[\sum_{t=1}^T\tilde{\ell}_t(i_t)-\sum_{t=1}^T\tilde{\ell}_t(i^*)]\nonumber\\
=&\mathbb{E}[\sum_{t=1}^Tp_t \cdot \tilde{\ell}_t-\sum_{t=1}^T {\tilde{\ell}}_t(i^*)]
\nonumber\\
\leq&\frac{\log |K|}{\eta} + \eta\sum_{t=1}^T\mathbb{E}[p_t\cdot \tilde{\ell}_t^2]\nonumber\\
\leq&\frac{\log |K|}{\eta} + \eta T 2^{H+1},\nonumber
\end{align}
which completes the proof.
\end{proof}
Next we prove Lemma \ref{lem:cost} in the main text. 
\begin{replemma}{lem:cost}
For sufficient large $T$, 
suppose for $m=1,\dots,N-1$ that the { parameters} $d_m$ of an MSET are chosen recursively,
\begin{align}
    &d_1 = \left \lfloor{-\log\left(\frac{1}{\sqrt{\lambda}}{\sum_{j=2}^{N-1}c_j}/{\sum_{j=1}^{N-1}c_j}\right)}\right \rfloor,\nonumber\\
    &d_m  =\left \lfloor{-\log\left(\frac{1}{\sqrt{\lambda}}{\sum_{j=m+1}^{N-1}c_j}/{\sum_{j=1}^{N-1}c_j}\right)}\right \rfloor -\sum_{n=1}^{m-1}d_{n},\nonumber\\
    &i=2,\dots,N-2,\nonumber\\
    &d_{N-1} = \log (T^{1/3}/\log |\mathcal{K}|) - \sum_{m=1}^{N-2}d_{m}.
\end{align}
Then \alg~results in cumulative costs
\begin{align}
    \mathbb{E}[\sum_{t=1}^T\Gamma^t]
    = \mathcal{O} \bigg( \sum\limits_{m=1}^{N-1}{\sqrt{\lambda}}c_mT^{2/3}\log |\mathcal{K}|\log \frac{T^{1/3}}{\log|\mathcal{K}|} \bigg). 
\end{align}
\end{replemma}

\begin{proof}
The proof follows by showing firstly that the movement cost is dominated by a HST metric, and secondly that under the tree metric the cumulative cost is bounded by the quantity in the lemma. To define the HST metric formally, let us introduce the following terminology in accordance to \citep{pmlr-v65-koren17a}. Given $u$, $v$ be nodes in the MSET $\mathcal{T}$, let LCA($u,v$) be their least common ancestor node. Then the scaled HST metric is defined as follows:
\begin{align}\label{HST}
    \Delta_{\cal{T}}(u,v) = ({\sqrt{\lambda}}\sum_{j=1}^{N-1}c_j)\frac{2^{\text{level}(\text{LCA}(u,v))}}{2^{\text{depth}(\cal{T})}},~\forall u,v\in \mathcal{K}.
\end{align}
Under this metric, the cost incurred from changing variables in the $i^{\rm th}$ module is 
\begin{align*}
    ({\sqrt{\lambda}}\sum_{j=1}^{N-1}c_j)\frac{2^{d_i+\cdots+d_{N-1}}}{2^{d_1+\cdots+d_{N-1}}}=\frac{\sum_{j=1}^{N-1}c_j}{2^{d1+\dots+d_{i-1}}}.
\end{align*}
Then the condition of dominance over the original cost  is, for $i=1,\dots,N-2$,
\begin{align*}
    &\frac{1}{\sqrt{\lambda}}\frac{\sum_{j=1}^{N-1}c_j}{2^{d1+\dots+d_{i-1}}}\geq \sum_{j=i}^{N-1}c_i,\nonumber\\
    \Rightarrow & d_1+\cdots+d_{i-1} \leq -\log\left(
    \frac{\sum_{j=i}^{N-1}c_j}{\sum_{j=1}^{N-1}c_j}\right) - \frac{\log \lambda}{2}.
\end{align*}
Rearrangements of these linear inequalities yield the solution for $d_1$ to $d_{N-2}$ as
\begin{align}\label{cond}
    &d_1 = \left\lfloor{-\log\left(
    \frac{\sum_{j=2}^{N-1}c_j}{\sum_{j=1}^{N-1}c_j}\right)}- \frac{\log \lambda}{2}\right\rfloor,\nonumber\\
    &d_i = \left\lfloor{-\log\left(
    \frac{\sum_{j=i+1}^{N-1}c_j}{\sum_{j=1}^{N-1}c_j}\right)}- \frac{\log \lambda}{2}\right\rfloor - \sum_{n=1}^{i-1}d_n,\nonumber\\
    &i = 2,\dots,N-2. 
\end{align}
Under the condition in Eq.~(\ref{cond}), the cost incurred from the HST metric Eq.~(\ref{HST}) is larger than our original cost. Hence, an upper bound for the cost incurred from the metric will also bound our cumulative cost.

Now we bound the cumulative cost under this HST metric.
Observe $i_t$ and $i_{t-1}$ belongs to the same subtree on level $h$ of the tree with probability at least $1-2^{h-H}$, therefore we have
\begin{align}\label{costeq}
    \mathbb{E}[\Delta_{\cal{T}}(i_t,i_{t-1})]&\leq \sum_{j=1}^{N-1}{\sqrt{\lambda}}c_j\sum_{h=0}^{H-1}2^{h-H} \cdot 2^{h-1} \nonumber\\
    &\leq \sum_{j=1}^{N-1}{\sqrt{\lambda}}c_j\frac{H}{2^{H+1}}.
\end{align}
On the other hand, the condition of $d_{N-1}=\mathcal{O}(T^{1/3}/log |K|) - d_1- \cdots -d_{N-2}$ admits a non-negative solution of $d_{N-1}$ for sufficient large $T$. This condition implies an upper bound on $H = d_1+\cdots+d_{N-1} = \mathcal{O}(\log(T^{1/3}\log |K|))$. Finally, combining this upper bound of $H$ with Eq.~(\ref{costeq}) completes the proof.
\end{proof}
Now we are in the last stage of proving Theorem \ref{mainthm}. 
\begin{reptheorem}{mainthm}
For $1\leq m\leq N-1$, let $D_m$ denote the dimension of ${\cal{X}}_m$ and suppose for all $t>0$, we set $\beta_t=\Theta(\sqrt{\gamma_{t-1}+\ln T})$,  $\eta=\Theta(T^{-2/3}\sum_{m=1}^{N-1}D_m\log(\frac{LT^{1/3}}{D_m\log T}))$, and 
we have an MSET with a uniform partition of each ${\mathcal{X}}_m$ with diameters
$r_m=\frac{D_m}{L}T^{-\frac{1}{3}}\log T$, where the depth parameters $d_m$ follows from Lemma \ref{lem:cost}. 
Then \alg~achieves the expected movement regret
\begin{align*}
   \mathbb{E}[ R^+]=\mathcal{O}(\lambda(\sum_{j=1}^{N-1} c_j\sum_{m=1}^{N-1}D_m T^{\frac{2}{3}}(\log T)^2) +\gamma_T\sqrt{T}).
\end{align*}
\end{reptheorem}
\vspace{-4mm}
\begin{proof}
We first bound the ordinary regret. Choose $\beta_t =  1+\sigma \sqrt{2\left(\gamma_{t-1}+1+\ln T\right)}$. Then, with probability $1-1/T$, we have
\begin{align}\label{der1}
R(T)=&\sum_{t=1}^Tf(x_t)-f^*\nonumber\\
\overset{(a)}{\leq}&  \sum_{t=1}^T \alpha^t(x_t)- \min_{x\in {\mathcal{X}}} \alpha^t(x) + 2\beta_t\sigma_{t-1}(x_t)\nonumber\\
\overset{(b)}{\leq}&\sum_{t=1}^T\bar{\ell}_{t,0}(i_t)- \bar{\ell}_{t,0}(i^*)+L\sum_{i=1}^{N-1}r_iT +\sum_{t=1}^T 2\beta_t\sigma_{t-1}(x_t),
\end{align}
where (a) follows from Lemma \ref{lem:beta} and (b) from the fact that $\bar{\ell}_{t,0}(i^*)=\min\limits_{\mathbf{z}\in \mathcal{Z}}{\alpha_t(\mathbf{x}_{1:j-1}^{t-1},\mathbf{z})}$ for some $\mathcal{Z}$ and that $f$ is $L$-Lipschitz.   

Note that when the above inequality fails it only contributes to cumulative regret in expectation by $1/T \times {\mathcal{O}}(T) = {\mathcal{O}}(1)$, so we can ignore this term in later calculation.

Now, taking expectation on both sides of Eq.~(\ref{der1}) yields 
\begin{align}
\mathbb{E}[R] \overset{(c)}{\leq} &~\mathbb{E}[\sum_{t=1}^T (\bar{\ell}_{t,0}(i_t) - \bar{\ell}_{t,0}(i^*))]+L\sum_{i=1}^{N-1}r_iT 
+\mathcal{O}(\gamma_T\sqrt{T})\nonumber\\
\overset{(d)}{=}&~\mathbb{E}[\sum_{t=1}^T ({\tilde{\ell}}_{t}(i_t) - {\tilde{\ell}}_{t}(i^*))]+L\sum_{i=1}^{N-1}r_iT 
+\mathcal{O}(\gamma_T\sqrt{T})\nonumber\\
\overset{(e)}{=}&~\mathcal{O}(\sqrt{2^HT\log|\mathcal{K}|} +L\sum_{i=1}^{N-1}r_iT +\gamma_T\sqrt{T}),\nonumber
\end{align}
where (c) follows from Lemma \ref{lem:sigsum}, (d) from that $\mathbb{E}[\ell_t]=\mathbb{E}[\bar{\ell}_{t,0} + \sum_{j=0}^{H-1}\bar{\ell}_{t,j}]=\mathbb{E}[\bar{\ell}_{t,0}] + \sum_{j=0}^{H-1}\mathbb{E}[\sigma_{t,j}]\mathbb{E}[\bar{\ell}_{t,j}]=\mathbb{E}[\bar{\ell}_{t,0}]$, and (e) from Lemma \ref{smblem} where $\frac{\log |K|}{\eta} +\eta T2^{H+1}= \mathcal{O}(\sqrt{2^{H+1}T\log |\mathcal{K}|})$ for $\eta = \sqrt{2^{-H}T^{-1}\log|\mathcal{K}|}$.

On the other hand, the cumulative movement cost by Lemma \ref{lem:cost} is
\begin{align}\label{cost2}
    \sum_{t=1}^T \Gamma^t = \mathcal{O}(\sum_{j=1}^{N-1}{\sqrt{\lambda}}c_j\frac{H}{2^H}T).
\end{align}
From Eq.~(\ref{cost2}), we plug in $H = \log(T^{1/3}/\log|\mathcal{K}|)$, $r_i = \Theta(\frac{D_i}{L}T^{-1/3}\log T)$ and $|\mathcal{K}|=\Theta(\prod_{i=1}^{N-1}1/r_i^{D_i})$. 

Then, we have
\begin{align*}
    \eta =&\sqrt{2^{-H}T^{-1}\log|\mathcal{K}|}\nonumber\\
    =&\Theta(T^{-2/3}\sum_{i=1}^{N-1}D_i\log(LT^{1/3}/D_i\log T)),
\end{align*}
and 
\begin{align*}
    &\mathbb{E}[R^+] = \mathbb{E}[R] + \mathbb{E}[\sum_{t=1}^T \sqrt{\lambda}\Gamma^t] \nonumber\\
    \leq & \mathcal{O}(\sum_{j=1}^{N-1}c_jT^{\frac{2}{3}}\log T\log |K|+L\sum_{i=1}^{N-1}r_iT +\gamma_T\sqrt{T})\nonumber\\
     \leq & \mathcal{O}(\sum_{j=1}^{N-1}c_j\sum_{i=1}^{N-1}D_i T^{\frac{2}{3}}(\log T)^2 +\gamma_T\sqrt{T}),
\end{align*}
 which completes the proof.
\end{proof}

\subsection{Bounds on the maximum mutual information for common kernels}\label{subsec:boundmui}
The following lists known bounds of the maximum mutual information for common kernels 
\citep{Srinivas:2010:GPO:3104322.3104451}:
\begin{itemize}
    \item $\gamma_t=O(D\log t)$,\quad for linear kernel.
    \item $\gamma_t=O((\log t)^{D+1})$, \quad for Squared Exponential kernel. 
    \item $\gamma_t=O(t^{\frac{D(D+1)}{2\nu+D(D+1)}}\log  t)$\quad for Mat\'ern kernels with $\nu>1$,
\end{itemize}
where $D$ is the dimension of input space.

\vspace{-2mm}
\section{Practical considerations and implementation details}
\vspace{-2mm}\label{supp:sec:prac}

\label{sec:practical}
\subsection{Details on model selection}
Below we detail the extensions we use in the experiments to improve the algorithm's performance.\\
 
\noindent\textbf{Restart with Epochs:} A plausible strategy is to refresh the arm-selection probability every $\tau$ iterations to escape from local optimum. In our implementation we choose $\tau = 25$ as the default value.
    
\noindent\textbf{Adaptive Resolution Increase:} In experiments, a simple extension allows {\alg} to discard the arms that have probability of selection being less than a threshold ($\tau=0.9$ in our implementation), and partition each remaining subset into $2$ subsets. We found that combining this with restart can accelerate the optimization in many cases.
   
\noindent\textbf{Update of Kernel:} We choose RBF kernel and M\'{a}tern class. As commonly found in practice, we update our kernel hyperparameters every $25$ iterations based on the maximum likelihood estimation.
    
\noindent\textbf{Aggressive learning rate:} 
    Our experiments show that constant learning rate $\eta =1$ usually outperforms the rate $\Theta(T^{-2/3})$ suggested by the theory.

\subsection{Further design of MSET and partition strategies}
\paragraph{Construction of MSET: }A crucial part of algorithm is in the construction of the MSET, which involves partitioning the variables in each module, and setting the depth parameters (${d_i}$'s). For a MSET with $|\cal{K}|$ leaves to choose from, \alg ~requires solving $|\cal{K}|$ local BO optimization problems per iteration. Hence initially, we partition each variable space of module to two subsets only, and abandon subsets when their arm selection probability $p_t$ is below some threshold. In our experiments, we always set it to be ${0.2}/{|\cal{K}|}$, where $|\cal{K}|$ denotes the number of leaves of MSET. After that, we further divide the remaining subsets again to increase the resolution. This procedure could be iterated upon further although we typically do not go beyond two stages of refinement. %Theorem $1$ proposes setting $d_i$ according to $c_i$, which are sometimes not known a priori however. Thus, 
%Integrating a tree-based method like the \textit{High Confidence Tree (HCT)} \citep{azar2014online} could be used and we look to integrate it into our algorithm in the future. 

In our implementation, we set the depth parameter to be $d_i = 1$ or $d_i \propto \log {\lambda}c_i$ when $c_i$ could be estimated in prior. Empirically, we found that the performance is quite robust when $d_i\leq 5$ for the different cost ratios in both synthetic and real experiments we tested. To avoid accumulating cost too fast in early stages of \alg, we record the number of times that variable changes in the first module and 
dynamically increase the first depth parameters $d_1$ by $1$ every $20$ iterations when the the number has increased beyond $5$ ($1/4$ of the cycle) during the period. In all experiements, we have found this simple add-on perform on par or better than fixing depth parameters through an entire run.

\paragraph{Partitioning method:}
Although \alg~achieves theoretical guarantees with a uniform partition, such a partition does not fully leverage the structure of the function. For this reason, the computational complexity can be very large for high-dimensional problems.
On the other hand, we observe from our experiments that simple bisection aligned with coordinates yields good performance on many synthetic data and on our neural data.
To further improve the performance, we adopt the multi-scale optimization strategy \citep{wang2014bayesian,azar2014online}, which adaptively increases the partition resolution through iterations. 
Practically it has often leads to more computational savings, which involves partitioning more finely in regions that have high rewards. A simple version of this strategy is also employed in our experiments where regions are discarded with probability of selection being below some threshold (typically $0.1$) and the remaining regions are further partitioned with increasing resolution. 
Another remedy is to use domain specific information to help restrict the search space or define the hierarchy in the MSET. The generality of the MSET makes it possible to use expert or prior knowledge to constrain switching between specific sets of variables that may be implausible. For instance, in our study of application in neuroscience, there are certain combinations of parameters that would violate certain size constraints related to the underlying biology that could be incorporated into the design of the MSET. 
 
\section{Further details on the neuroimaging experiments}
\vspace{-2mm}

\begin{figure}[t!]
\centering
\includegraphics[width=0.8\textwidth]
       {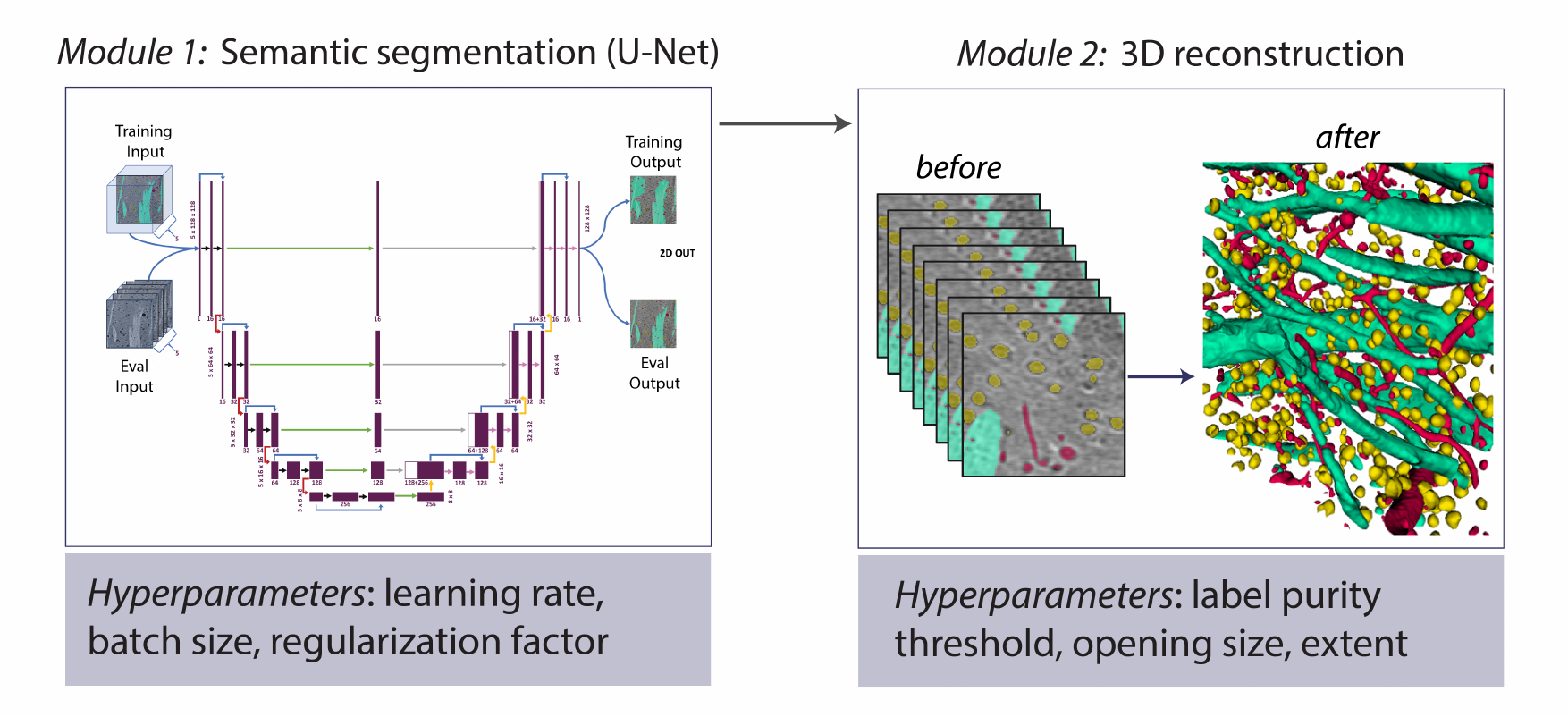}
     
\caption{\footnotesize {\em Pipeline for neuroimage segmentation.} From from left to right, we show the training of U-Net which outputs segmentation of 2D images, and a post-processed 3D reconstruction.
\label{fig:neuropipeline}}
\end{figure}

\begin{figure}[t!]
         \centering \hspace{10mm}
     {\includegraphics[width=0.7\textwidth]{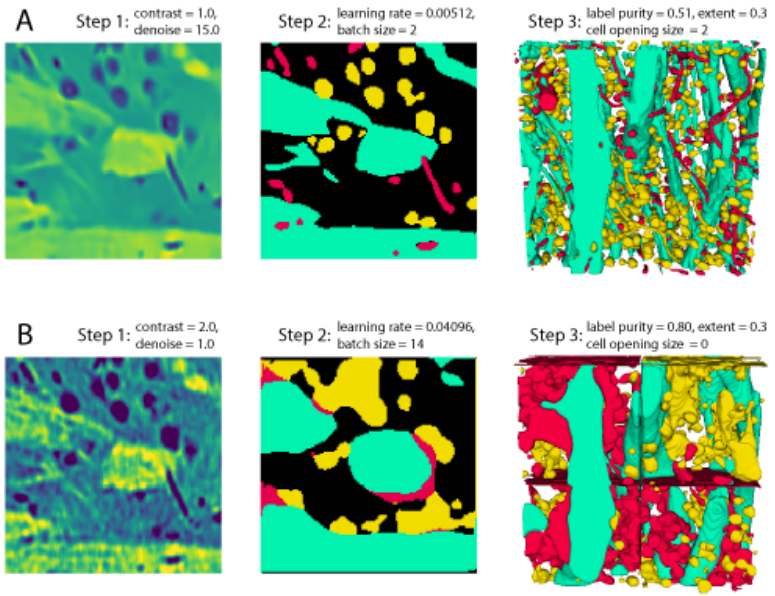}}
     \newline
     {\includegraphics[width=0.55\textwidth]{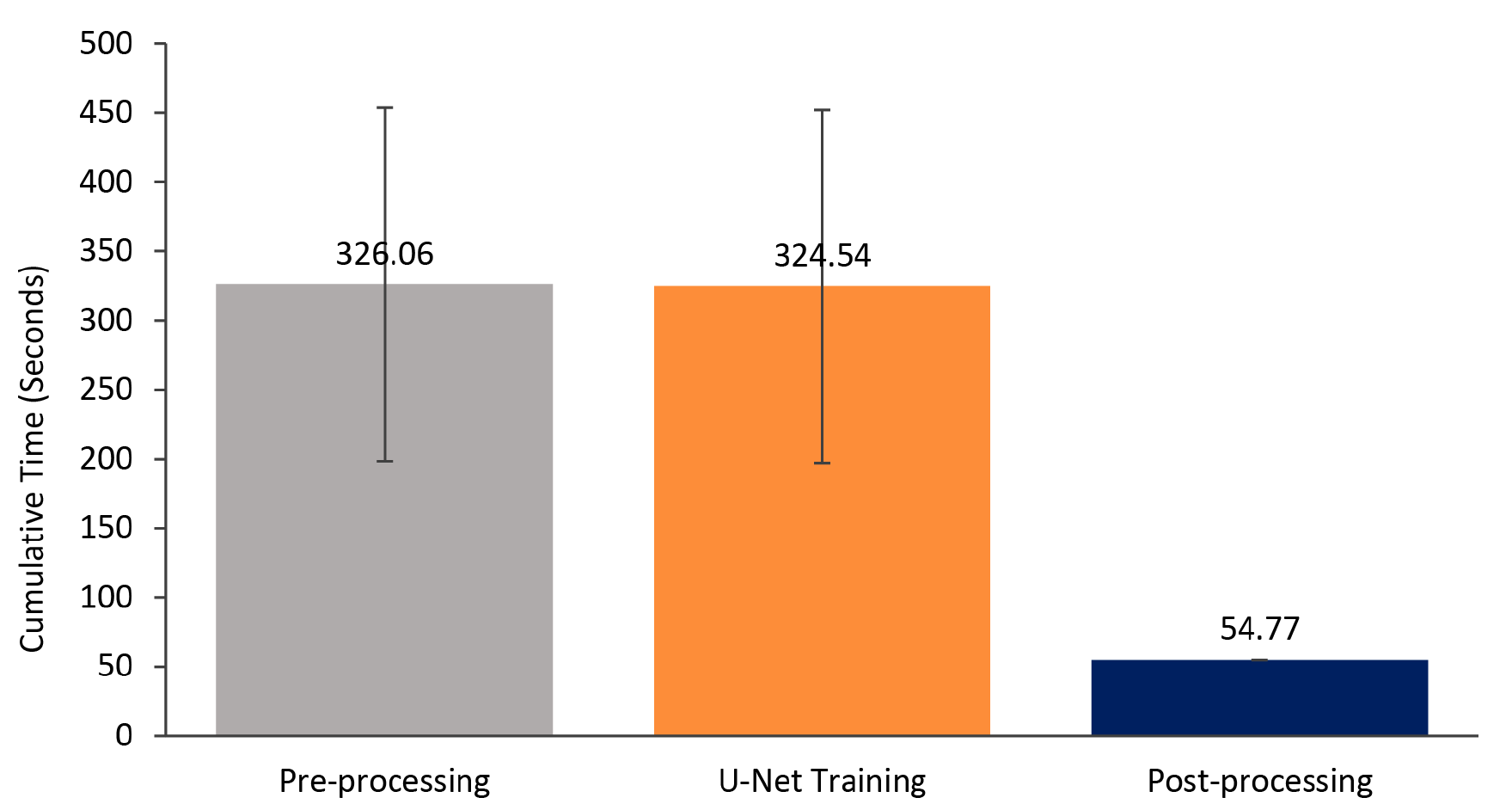}}

\caption{\label{fig:recons}\footnotesize {\em Results from the three-stage pipeline for the optimal set of parameters and a suboptimal set.} 
%The f1-score measured in step 3 is $77.129\%$. 
Along the top row (A), we show the results obtained for an optimal set of hyperparameters selected by our approach (as measured by the f1-score). Along the bottom row (B), we show the same results for a suboptimal hyperparameter combination with poor performance. Below, we show the statistics of the timing costs for each stage of the pipeline.
%The mean and standard deviation across 100 random hyperparameters was computed for each module, $326.059 \pm 128.390$ sec (pre), $324.539 \pm 128.367$ sec (U-net), and $54.773 \pm 0.894$ sec (post).
%The f1-score measured in step 3 is only $27.692\%$.
%The hyperparameters used in each step are indicated along the top of each row of images, and the average f1-score achieved by that combination is indicated under each row. 
}

\label{fig:image}
%\vspace{-6mm}
\end{figure}
Figure \ref{fig:neuropipeline} illustrates the scheme of our brain-imaging experiment without pre-processing.
In this brain mapping pipeline, we varied the U-Net training hyperparameters and the 3D reconstruction post-processing hyperparameters. In the first module (U-Net training), we optimized the learning rate $\in[1\times 10^{-7}, 1\times 10^{-1}]$ and batch size $\in[4,12]$ for the U-Net. In the second module, we applied post-processing operations to the U-Net output 3D reconstructions, including label purity $\in[0.51,0.8]$, cell opening size $\in[0,2]$, and a shape parameter (extent) to determine whether uncertain components are either cells or blood vessels $\in[0.3,0.8]$. 
%Our grid search of hyperparameter combinations consisted of 101 values for learning rate ($[1\times 10^{-7}$ and then equal spacing between $1\times 10^{-3}$ and $1\times 10^{-1}$), 5 values for batch size (equal spacing), 20 values for label purity (equal spacing), 3 values for cell opening size (equal spacing), and 20 values for extent (equal spacing). Thus, in the 2-module experiment, we created 505 U-Net outputs, and 606,000 final hyperparameter combinations scores to test \alg's performance. 

We also performed a 3-module experiment by adding a pre-processing before U-Net training. We varied the pre-processing hyperparameters, U-Net training hyperparameters, and 3D reconstruction post-processing hyperparameters. In the pre-processing, we used a contrast parameter $\in [1,2]$ and denoising parameter  $[1,15]$ (regularization strength in Non-Local Means \citep{buades2011non}), and in the second module (U-Net training), we varied the learning rate $\in[1\times 10^{-5}, 8.192\times 10^{-2}]$ and batch size $\in[2,14]$. During the third module (post-processing of 3D reconstructions), we varied label purity $\in[0.51,0.8]$, cell opening size $\in[0,2]$, and extent $\in[0.3,0.8]$.
In our experiments, we define the cost to be the aggregate recorded clock time for generating an output after changing a variable in a specific module (see Figure \ref{fig:recons}, right). To test \alg~on the problem, we gathered a data set consisting of $606,000$ combinations of hyperparameters by exhaustive search.

\section{Pseudo-code for the Slowing Moving Bandit Algorithm}
\label{supp:smb}
For completeness, we include a pseudo-code for slowly moving bandit algorithm below.
\begin{algorithm}[H]
\begin{algorithmic}[1]
\caption{~Slowly Moving Bandit (SMB) }\label{smbalg}
\STATE {Input: }A tree $\mathcal{T}$ with a set of finite leaves $K, \eta>0$.
\STATE {Initialize: } $p_{1}=\text{Unif}(K), h_{0}=H$ and $i_{0} \sim p_{1}$
\vspace{0.5mm}
 \FOR {$t=1$ to $T$}
\STATE Select arm $i_t\sim p_t(\cdot|A_{h_{t-1}}(i_{t-1}))$.
\vspace{0.5mm}
\STATE Let $\sigma_{t,h}$, $h=1,\dots,H-1$, be i.i.d. Unif($\{-1,1\}$). 
\vspace{0.5mm}
\STATE let $h_t=\min\{0\leq h\leq H:\sigma_{t,h}=-1\}$ where $\sigma_{t,H}=-1$. 
\STATE Compute vectors $\bar{\ell}_{t,0},\dots,\bar{\ell}_{t,H-1}$ recursively via 
$\bar{\ell}_{t,0}(i) = \frac{{\bf 1} (i_t=i)}{p_t(i)}\ell_t(t)$, and for all $h\geq 1$: \\
 $$\bar{\ell}_{t,h}(i)=-\frac{1}{\eta}\log\left(\sum_{j\in A_h(i)}\frac{p_t(j)\zeta_{t,h}(j)}{p_t(A_h(i))}\right),~~
\zeta_{t,h}(j) = e^{-\eta(1+\sigma_{t,h-1})\bar{\ell}_{t,h-1}(j)}.$$\\
\STATE $\tilde{\ell}_t=\bar{\ell}_{t,h}+\sum_{h=0}^{H-1}\sigma_{t,h}\bar{\ell}_{t,h}.$ 
\vspace{1mm}
\STATE $p_{t+1}=\frac{p_t(i)e^{-\eta{\ell}_t(i)}}{\sum_{j=1}^{|\mathcal{K}|}p_t(j)e^{-\eta{\ell}_t(j)}},~\forall i\in \mathcal{K}.$ 
\vspace{1mm}
\ENDFOR
\end{algorithmic}
\end{algorithm}
\vspace{-3mm}

 \section{\textbf{Further Experiments on Synthetic Functions}
}
\label{supp:sec:syn}
\begin{figure*}[ht!]
\centering

   \includegraphics[width=\textwidth]
   {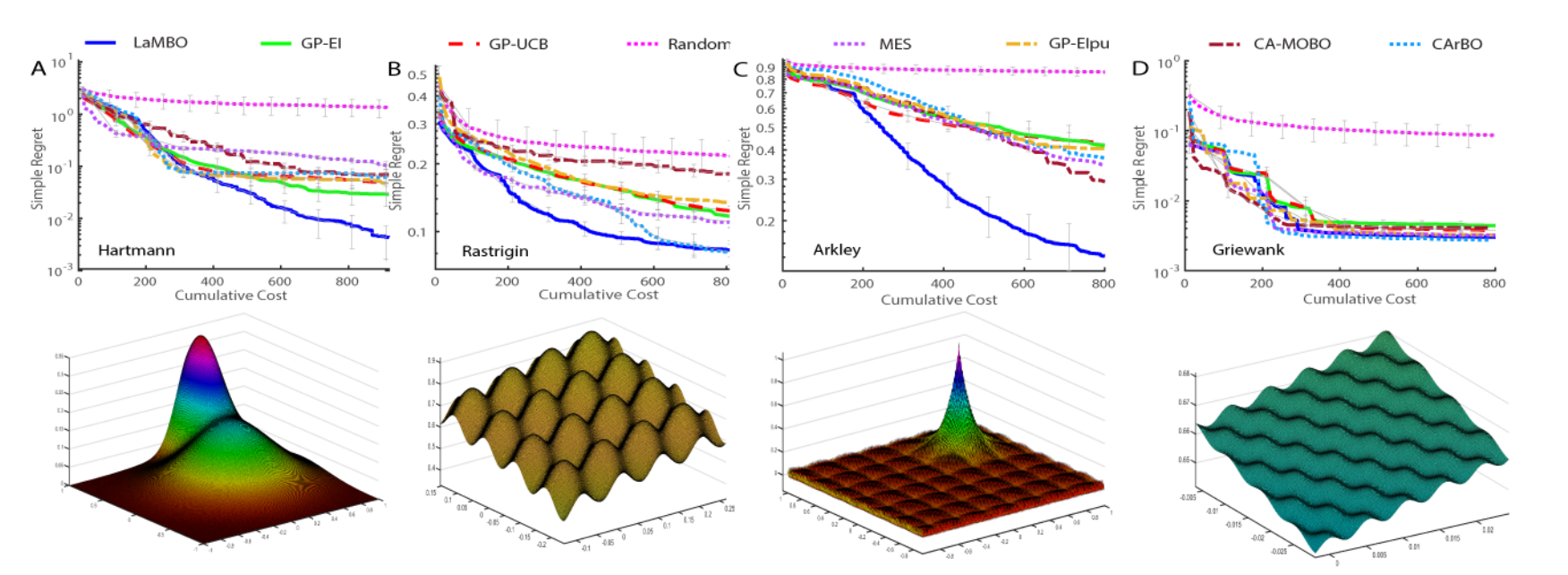}

\vspace{3mm}
\caption{\footnotesize {\em Synthetic functions.} 
This section contains details and further experiment results on the synthetic functions.
We compare 
\alg~with other BO algorithms on four synthetic functions, (A) Hartmann 6D, (B) Rastrigin 6D, (C) Ackley 8D, and (D)  Griewank 6D. The plots on the top shows the regret performance and the 3D plots on the bottom show their surface. 
   }
\label{fig:synsupp}
\end{figure*}

The synthetic functions used in the experiment are taken from \citep{simulationlib} and \citep{kandasamy2016gaussian}. We use linear transformation to normalize all the function to the range $[0,1]$.

We compare 
\alg~with other BO algorithms on four synthetic functions, (A) Hartmann 6D, (B) Rastrigin 6D, (C) Ackley 8D, and (D)  Griewank 6D. The plots on the top shows the regret performance, the plots on the button show their surface. We observe when objective have multiple local optimum comparable with the global one, \alg~has comparable performance with the alternative. However, \alg~performs significantly better than the baselines when the objective has a sharper global optimum. Unlike deterministic decision rule proposed in the alternatives, \alg~has randomized decision rule and does not rely on the GP regression alone, which allows it to have more incentive for exploration.
%The plots on the top shows the regret performance, the plots on the button show their surface. We observe when objective have multiple local optimum comparable with the global one, \alg~has comparable performance with the alternative. However, \alg~performs significantly better than the baselines when the objective has a sharper global optimum. Unlike deterministic decision rule proposed in the alternatives, \alg~has randomized decision rule and does not rely on the GP regression alone, which allows it to have more incentive for exploration.
%Below we provide further details on the synthetic function used in the experiments.

\noindent\textbf{Hartmann 6D function:}\\
The function is $f(x) = \sum_{i=1}^4\alpha_i\exp(-\sum_{j=1}^6{A_{ij}(x_j-P_{ij})})$, where 
 \begin{align*}
\alpha&=[1,1,2,3,3.2],\\
     A&=\left[\begin{array}{ccccc}
{10} & {3} & {17} & {3.5} & {1.7}  \\
{0.05} & {10} & {17} & {0.1} & {8} \\
{3} & {3.5} & {1.7} & {10} & {17} \\
{17} & {8} & {0.05} & {10} & {0.1}
\end{array}\right], \nonumber\\ P&=10^{-4} \times\left[\begin{array}{ccccccc}
{1312} & {1696} & {5569} & {124} & {8283} & {5886} \\
{0.05} & {10} & {17} & {8} & {17} & {8} \\
{17} & {8} & {0.05} & {10} & {0.1} & {14}
\end{array}\right],
 \end{align*}  
 and the domain is $[0,1]^6$.\\
 
 \noindent\textbf{Ackley 8D function:}
 \begin{align*}
 &f(\mathbf{x})
-20 \exp (-0.2 \sqrt{\frac{1}{8} \sum_{i=1}^{8} x_{i}^{2}})-\exp \left(\frac{1}{8} \sum_{i=1}^{8} \cos \left(2\pi x_{i}\right)\right)\\
 +&20+\exp (1),
\end{align*}
where the domain is $[-32.768,-32.768]^8$.\\

\noindent\textbf{Rastrigin 6D function:}
\begin{align*}
    f(\mathbf{x})=60+\sum_{i=1}^{6}\left[x_{i}^{2}-10 \cos \left(2 \pi x_{i}\right)\right],
\end{align*}
where the domain is $[-5.12,5.12]^6$.\\

\noindent\textbf{Griewank 6D function:}
\begin{align*}
    f(\mathbf{x})=\sum_{i=1}^{6} \frac{x_{i}^{2}}{4000}-\prod_{i=1}^{6} \cos \left(\frac{x_{i}}{\sqrt{i}}\right)+1,
\end{align*}
where the domain is $[-600,600]^6$.

$$\alpha_c(\mathbf{x}) = \alpha(\mathbf{x}) / c(\mathbf{x})$$

\end{document}